\documentclass{article}

\PassOptionsToPackage{numbers, compress}{natbib}
\usepackage[final]{neurips_2022}

\usepackage[utf8]{inputenc} 
\usepackage[T1]{fontenc}    
\usepackage[breaklinks=true,letterpaper=true,colorlinks,bookmarks=false]{hyperref}       
\usepackage{url}            
\usepackage{booktabs}       
\usepackage{amsfonts}       
\usepackage{nicefrac}       
\usepackage{microtype}      
\usepackage{xcolor}         
\usepackage[pdftex]{graphicx}

\usepackage{colortbl}
\usepackage{amsmath}
\usepackage{pifont}
\usepackage{bbm}
\usepackage{adjustbox}
\usepackage{multirow}
\usepackage{wrapfig}
\usepackage{tablefootnote}
\usepackage{afterpage}
\newcommand{\etc}{{\it etc}}
\newcommand{\ie}{{\it i.e.}}
\newcommand{\eg}{{\it e.g.}}

\newcommand{\HF}[1]{\textcolor[rgb]{1.00,0.00,0.00}{#1}}
\newcommand{\tabincell}[2]{\begin{tabular}{@{}#1@{}}#2\end{tabular}}

\definecolor{mygray}{gray}{.9}

\usepackage{pgfplots}
\DeclareUnicodeCharacter{2212}{−}
\usepgfplotslibrary{groupplots,dateplot}
\usetikzlibrary{patterns,shapes.arrows}
\pgfplotsset{compat=newest}

\usepackage{subfig}

\title{SwinTrack: A Simple and Strong Baseline for Transformer Tracking}

\author{
Liting Lin$^{1,2}$\thanks{Equal Contributions.} \hspace{9pt}
Heng Fan$^3$\footnotemark[1] \hspace{9pt}
Zhipeng Zhang$^4$\hspace{9pt} 
Yong Xu$^{1,2}$\hspace{9pt} 
Haibin Ling$^5$\\
$^1$School of Computer Science \& Engineering, South China Univ. of Tech., Guangzhou, China\\
$^2$Peng Cheng Laboratory, Shenzhen, China\\
$^3$Department of Computer Science and Engineering, University of North Texas, Denton, USA\\
$^4$DiDi Chuxing, Beijing, China\\
$^5$Department of Computer Science, Stony Brook University, Stony Brook, USA\\
l.lt@mail.scut.edu.cn, heng.fan@unt.edu, zhipeng.zhang.cv@outlook.com\\
yxu@scut.edu.cn, hling@cs.stonybrook.edu
}

\begin{document}

\maketitle

\begin{abstract}
  
  Recently Transformer has been largely explored in tracking and shown state-of-the-art (SOTA) performance. However, existing efforts mainly focus on fusing and enhancing features generated by convolutional neural networks (CNNs). The potential of Transformer in representation learning remains under-explored. In this paper, we aim to further unleash the power of Transformer by proposing a simple yet efficient fully-attentional tracker, dubbed \textbf{SwinTrack}, within classic Siamese framework. In particular, both representation learning and feature fusion in SwinTrack leverage the Transformer architecture, enabling better feature interactions for tracking than pure CNN or hybrid CNN-Transformer frameworks. Besides, to further enhance robustness, we present a novel motion token that embeds historical target trajectory to improve tracking by providing temporal context. Our motion token is lightweight with negligible computation but brings clear gains. In our thorough experiments, SwinTrack exceeds existing approaches on multiple benchmarks. Particularly, on the challenging LaSOT, SwinTrack sets a new record with \textbf{0.713} SUC score. It also achieves SOTA results on other benchmarks. We expect SwinTrack to serve as a solid baseline for Transformer tracking and facilitate future research. Our codes and results are released at \url{https://github.com/LitingLin/SwinTrack}.

\end{abstract}

\section{Introduction}

\begin{figure}
\centering
\includegraphics[width=0.8\linewidth]{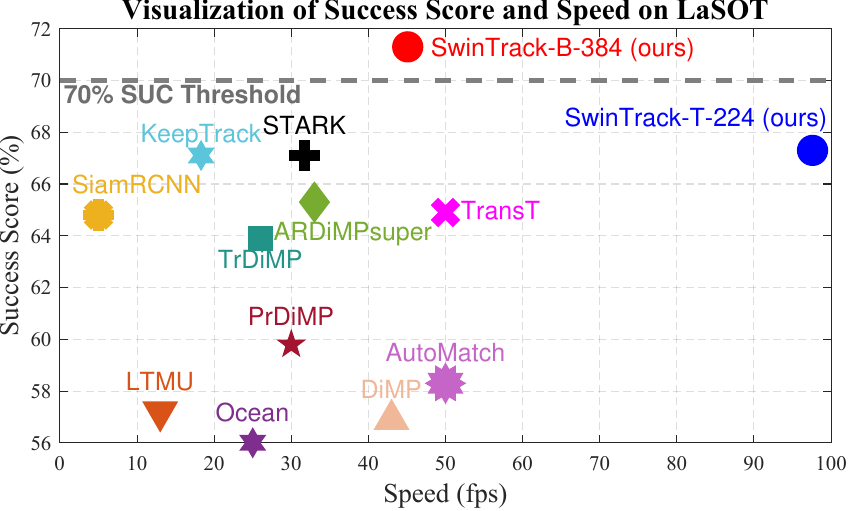}
\caption{
Comparison on LaSOT~\cite{fan2019lasot}. Our tracker (SwinTrack-B-384) sets a new record with 0.713 SUC score and still runs efficiently at around 45 {\it fps}. A lighter version (SwinTrack-T-224) achieves 0.672 SUC score and runs at around 96 {\it fps}, which is on par with existing SOTAs in accuracy but much faster. Best viewed in color for all figures.
}
\label{auc_fps_fig}
\end{figure}


Visual tracking has seen considerable progress since deep learning. In particular, the recent Transformer~\cite{vaswani2017attention} has significantly pushed the state-of-the-art in tracking owing to its ability in modeling long-range dependencies. However, existing methods usually leverage Transformer for fusing and enhancing features generated from convolutional neural networks (CNNs), \eg, ResNet~\cite{he2016deep}. The potential of exploiting Transformer for feature representation learning is largely under-explored.

Recently, Vision Transformer (ViT)~\cite{dosovitskiy2020imagevit} has exhibited great potential in robust feature representation learning. Particularly, its extension Swin Transformer~\cite{liu2021Swin} has achieved state-of-the-art (SOTA) results on multiple tasks. Taking inspiration from this, we argue, besides the feature fusion, the representation learning in tracking can also benefit from Transformer via attention. Thus motivated, we propose to develop a fully attentional tracking framework based on Siamese architecture. Specifically, both the feature representation learning and the feature fusion of template and search region are realized by Transformer. More concretely, we borrow the architecture of the powerful Swin Transformer~\cite{liu2021Swin} and adapt it to Siamese tracking. Note that, other Transformer architectures can be used. For feature fusion, we introduce a simple homogeneous concatenation-based fusion architecture, without a query-based decoder.

Moreover, taking into consideration that tracking is a temporal task, we propose a novel motion token to improve robustness. Inspired by that the target usually moves smoothly in a short period, motion token is represented by the historical target trajectory within a local temporal window. We incorporate the (single) motion token in the decoder of feature fusion to leverage motion information during tracking. Despite being conceptually simple, our motion token can effectively boost tracking performance, with negligible computation.

\begin{figure*}
    \centering
    \includegraphics[width=\linewidth]{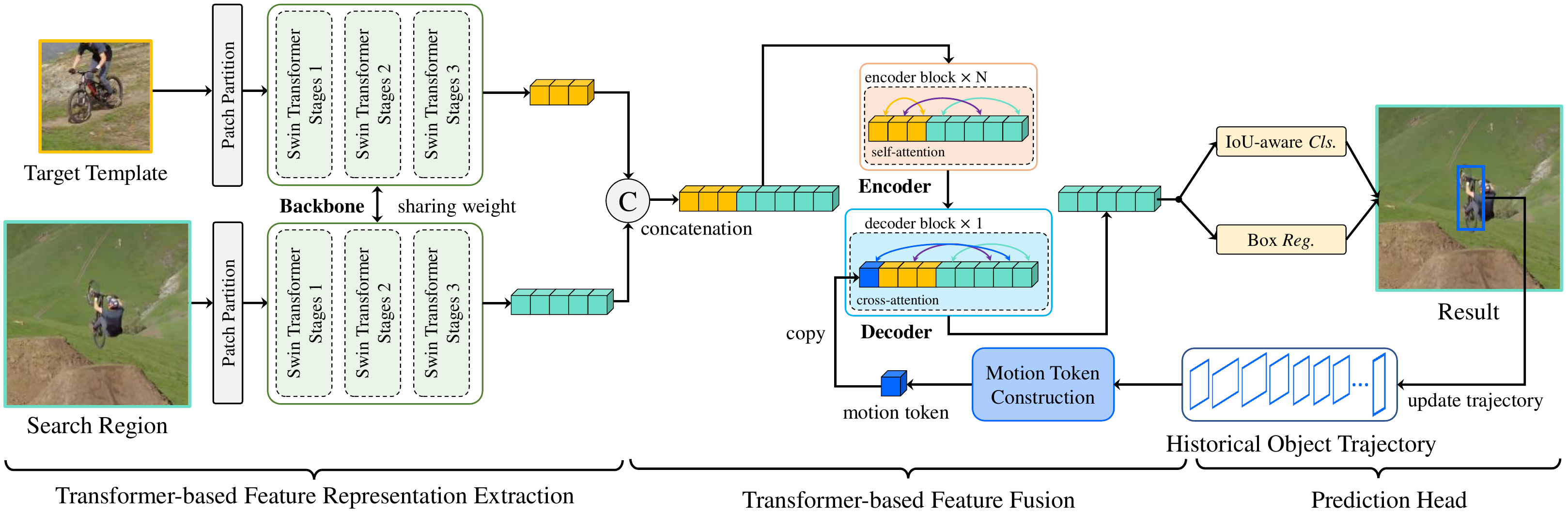}
    \caption{Architecture of SwinTrack, which contains three parts including Transformer-based feature representation extraction, Transformer-based feature fusion and prediction head. Our SwinTrack is a simple and neat tracking framework without complex designs such as multi-scale features or temporal template updating, yet demonstrating state-of-the-art performance. {\it Best viewed in color}.}
    \vspace{-1em}
    \label{overview}
\end{figure*}

We name our framework SwinTrack. As a pure Transformer framework, SwinTrack enables better interactions inside the feature learning of template and search region and their fusion compared to pure CNN-based~\cite{bertinetto2016siamfc,li2018highsiamrpn} and hybrid CNN-Transformer~\cite{TransT,wang2021transformer,stark} frameworks, leading to more robust performance (see Fig.~\ref{auc_fps_fig}). Fig.~\ref{overview} demonstrates the architecture of SwinTrack. We conduct extensive experiments on five large-scale benchmarks to verify the effectiveness of SwinTrack, including LaSOT~\cite{fan2019lasot}, LaSOT$_{\mathrm{ext}}$~\cite{fan2021lasot}, TrackingNet~\cite{muller2018trackingnet}, GOT-10k~\cite{Huang2021got10k} and TNL2k~\cite{wang2021tnl2k}. On all benchmarks, SwinTrack achieves promising results and meanwhile runs fast at 45 {\it fps}. In particular, on the challenging LaSOT, SwinTrack sets a new record of 71.3 SUC score, surpassing the strongest prior tracker~\cite{stark} (to date) by 3.1 absolute percentage points and crossing the 0.7 SUC threshold {\it for the first time} (see Fig.~\ref{auc_fps_fig} again). It also achieves 49.1 SUC, 84.0 SUC, 72.4 AO and 55.9 SUC scores on LaSOT$_{\mathrm{ext}}$, TrackingNet, GOT-10k and TNL2k respectively, which are better than or on par with state-of-the-arts (SoTAs). In addition, we provide a lighter version of SwinTrack that obtains comparable results to SoTAs but runs much faster at around 98 {\it fps}. 

In summary, our contributions are as follows: ({\bf i}) We propose SwinTrack, a simple and strong baseline for fully attentional tracking; ({\bf ii}) We present a simple yet effective motion token, enabling the integration of rich motion context during tracking, further boosting the robustness of SwinTrack, with negligible computation; ({\bf iii}) Our proposed SwinTrack achieves state-of-the-art performance on multiple benchmarks. We believe SwinTrack further shows the potential of Transformer and expect it to serve as a baseline for future research.

\section{Related Work}

{\bf Siamese Tracking.} The Siamese tracking methods formulate tracking as a matching problem and aim to offline learn a generic matching function for this task. The seminal method of~\cite{bertinetto2016siamfc} introduces a fully convolutional Siamese network for tracking and shows a good balance between accuracy and speed. To improve Siamese tracking in handling scale variation, the work of~\cite{li2018highsiamrpn} incorporates the region proposal network (RPN)~\cite{ren2015faster} into the Siamese network and proposes the anchor-based tracker, showing higher accuracy with faster speed. Later, numerous extensions have been presented to improve Siamese tracking, including deeper backbone~\cite{li2019evolutionsiamrpnp}, multi-stage architecture~\cite{Fan_2019_CVPR,fan2020cract}, anchor-free Siamese trackers~\cite{Ocean_2020_ECCV}, deformable attention~\cite{yu2020deformable}, to name a few.

{\bf Transformer in Vision.} Transformer~\cite{vaswani2017attention} originates from natural language processing (NLP) for machine translation and has been introduced to vision recently and shows great potential. The work of~\cite{carion2020enddetr} first uses Transformer for object detection and achieved promising results. To explore the capability of Transformer in representation learning, the work of~\cite{dosovitskiy2020imagevit} applies Transformer to construct backbone network, and the resulting Vision Transformer (ViT) attains excellent performance compared to convolutional networks while requiring fewer training resources, which encourages many extensions upon ViT\cite{touvron2021training,chen2021crossvit,yuan2021tokens,wang2021pyramid,liu2021Swin}. Among them, the Swin Transformer~\cite{liu2021Swin} has received extensive attention. It proposes a simple shifted window strategy to replace the fixed-patch method in ViT, which significantly improves efficiency and meanwhile demonstrates state-of-the-art results on multiple image tasks. Our work is inspired by Swin Transformer, but differently, we focus on the video task of visual tracking.

{\bf Transformer in Tracking.} Inspired by the success in other fields, researchers have leveraged Transformer for tracking. The method of~\cite{TransT} applies Transformer to enhance and fuse features in the Siamese tracking for improvement. The approach of~\cite{wang2021transformer} uses Transformer to exploit temporal features to improve tracking robustness. The work of~\cite{stark} introduces a new transformer architecture dedicated to visual tracking, explores the Spatio-temporal Transformer by integrating the model updating operations into a Transformer module.

Our SwinTrack is related to but significantly different from the above Transformer-based trackers. Specifically, the aforementioned methods mainly apply Transformer to fuse convolutional features and belong to the hybrid CNN-Transformer architecture. Unlike them, SwinTrack is a pure Transformer-based tracking architecture where both representation learning and feature fusion are realized with Transformer, enabling the exploration of better features for robust tracking.

\section{Tracking via Vision-Motion Transformer}

We present SwinTrack, a vision-motion integrated Transformer for object tracking, in Fig.~\ref{overview}.  The proposed framework contains three main components, \ie, the Swin-Transformer backbone for feature extraction,  the encoder-decoder network for mixing vision-motion cues, and the head network for localizing targets. In the following sections, we first shortly describe the Swin backbone network, then elaborate on the proposed vision-motion encoder-decoder. Afterward, we give a discussion about our method and shortly describe the network head and training loss.

\subsection{Swin-Transformer for Feature Extraction}

The deep convolutional neural network has significantly improved the performance of trackers. Along with the advancement of trackers, the backbone network has evolved twice: AlexNet~\cite{krizhevsky2012imagenet} and ResNet~\cite{he2016deep}. Swin-Transformer~\cite{liu2021Swin}, in comparison to ResNet, can give a more compact feature representation and richer semantic information to assist succeeding networks in better localizing the target objects (demonstrate in the ablation study demonstrated in the ablation study), which is thus chosen for basic feature extraction in our model.

Our tracker, following Siamese tracking framework~\cite{bertinetto2016siamfc}, requires a pair of image patches as inputs, \ie, template image $\mathbf{z}\in\mathbb{R}^{H_z\times W_z\times 3}$ and search region image $\mathbf{x}\in\mathbb{R}^{H_x\times W_x\times 3}$. As in the typical Swin-Transformer procedure, template and search region images are divided to non-overlapped patches and sent to the network, which generates template tokens (dubbed \textbf{T-tokens}) $\varphi(\mathtt{z})\in\mathbb{R}^{\frac{H_z}{s}\frac{W_z}{s}\times C}$ and search region tokens (dubbed \textbf{S-tokens}) $\varphi(\mathtt{x})\in\mathbb{R}^{\frac{H_x}{s}\frac{W_x}{s}\times C}$. $s$ is the stride of the backbone network. Since there is no dimension projection in our model, $C$ is the hidden dimension of the whole model.

\subsection{Vision-Motion Representation Learning}
The essential step for matching-based visual tracking is injecting the template information into the search region. In our framework, we adopt an encoder to fuse the features from the template and the search region, meanwhile, a decoder is arranged to achieve vision-motion representation learning, as illustrated in Fig.~\ref{overview}.

{\bf Encoder for fusing template and search tokens.} The encoder contains a sequence of Transformer blocks where each consists of a multi-head self-attention (MSA) module and a feed-forward network (FFN). FFN contains a two-layers multi-layer perceptron (MLP), GELU activation layer is inserted after the first linear layer. Layer normalization (LN) is always conducted before every module (MSA and FFN). Residual connection is applied to MSA and FFN modules.

Before feeding the features into the encoder, the template and search region tokens are concatenated along spatial dimensions to generate a mixing representation $\mathbf{f}_m$. For each block, the MSA module computes self-attention over mixing union representation, which equals to separately conducting self-attention on T-tokens/S-tokens and meanwhile performing cross-attention between T-tokens and S-tokens, but more efficient. FFN refines the features generated by MSA. When the tokens get out of the encoder, a de-concatenation operation is arranged to decouple the template and search region tokens. The process of encoder can be expressed as:
\begin{equation}
\begin{split}
\mathbf{f}_{m}^1 & = {\rm Concat}(\varphi(\mathbf{z}),\varphi(\mathbf{x})) \\
&\dots \\
\mathbf{f}_{m}^{l'} & = \mathbf{f}_{m}^l+{\rm MSA}({\rm LN}(\mathbf{f}_{m}^l)) \\
\mathbf{f}_{m}^{l+1} & = \mathbf{f}_{m}^{l'}+{\rm FFN}({\rm LN}(\mathbf{f}_{m}^{l'})) \\
&\dots \\
\mathbf{f}_{z}^{L},\mathbf{f}_{x}^{L} & = {\rm DeConcat}(\mathbf{f}^L),
\end{split}
\end{equation}
where $l$ denotes the $l$-th layer and $L$ denotes the number of blocks.

{\bf Decoder for fusing vision and motion information.} 
Before describing the architecture of decoder, we first detail how to generate a motion token (dubbed \textbf{M-token}). Motion token is the embedding of the historical trajectory of the target object. The past object trajectory is represented as a set of target object box coordinates, $T=\{ \mathbbm{o}_1, \mathbbm{o}_2, ...,  \mathbbm{o}_t \} $, where $t$ represents the frame index, $\mathbbm{o}$ is the bounding box of target object. $\mathbbm{o}$ is defined by the top-left and bottom-right corners of the target object, denotes as $\mathbbm{o}_{t} = (o^{x_1}_t, o^{y_1}_t, o^{x_2}_t, o^{y_2}_t)$. For flexible modeling, a sampling process is required to ensure the following properties: \textbf{1) fixed length}, \textbf{2) focusing on the latest trajectories} and \textbf{3) reducing redundancy}. In our method, we sample object trajectory as:

\begin{equation}
\begin{split}
\mathcal{T} &= \{ \mathbbm{o}_{s(1)}, \mathbbm{o}_{s(2)}, ..., \mathbbm{o}_{s(n)} \}, \;\;
{\rm where}\ s(i) = max(t - i \times \Delta, 1),
\end{split}
\end{equation}

$n$ is the number of sampled object trajectories, $\Delta$ is the fixed sampling interval. For Siamese tracker, the search region is cropped from the input image. In detail, a cropping with resizing operation can be used to describe the process. Giving the point in the input image as $(\mathbf{x}_i, \mathbf{y}_i)$, the corresponding point in the search region as $(\mathbf{x}_o, \mathbf{y}_o)$, we can formulate the cropping process employed in pre-processing of the Siamese Tracker as $\mathbf{x}_o = (\mathbf{x}_i - i_x) s_x + o_x \  \textrm{and} \  \mathbf{y}_o = (\mathbf{y}_i - i_y) s_y + o_y$,
where $(i_x, i_y)$ is the center of the cropping window in the input image, $(s_x, s_y)$ is the scaling factor, $(o_x, o_y)$ is the center of cropped and scaled window in the search region. We apply the same transformation on the sampled object trajectory to make the object trajectory invariant to the cropping, denoting $\Bar{\mathcal{T}} = \{ \Bar{\mathbbm{o}}_{s(1)}, \Bar{\mathbbm{o}}_{s(2)}, ..., \Bar{\mathbbm{o}}_{s(n)} \}$ as the result.

Then, to embed the transformed object trajectory into the network, we adopt four embedding matrices to embed the elements in box coordinates separately. We denotes the embedding matrix as $W\in{\mathbb{R}^{(\mathbbm{g}+1)\times{d}}}$, $\mathbbm{g}$ controls the embedding granularity of the object trajectory, $d$ is the size of each embedding vector. The last entry of the embedding matrix is used as the padding vector, indicating an invalid state, like object absence or out of the search region. Thus, we normalize the sampled target object box coordinates in the range $[1, \mathbbm{g}]$, and quantize to integers to get the index of embedding vector:

\begin{equation}
\begin{split}
\hat{\mathcal{T}} &= \{ \hat{\mathbbm{o}}_{s(1)}, \hat{\mathbbm{o}}_{s(2)}, ..., \hat{\mathbbm{o}}_{s(n)} \} , \\
{\rm where}\ \hat{\mathbbm{o}}_{s(i)} &= [ \mathbbm{n}(\Bar{\mathbbm{o}}_{s(i)}^{x_1}, w), \mathbbm{n}(\Bar{\mathbbm{o}}_{s(i)}^{y_1}, h), \mathbbm{n}(\Bar{\mathbbm{o}}_{s(i)}^{x_2}, w), \mathbbm{n}(\Bar{\mathbbm{o}}_{s(i)}^{y_2}, h) ], \\
\mathbbm{n}(o, l) &= \left\{
    \begin{array}{ll}
        \lfloor \frac{o}{l} \times \mathbbm{g} \rfloor  & \textrm{if} \ \textrm{valid}, \\
        \mathbbm{g} +1 & \textrm{else},
    \end{array}\right.
\end{split}
\end{equation}
where $(w, h)$ is the size of search region feature map.

Finally, the motion token $\mathbf{E}_{motion} \in \mathbb{R}^{1 \times d}$ is given by a concatenation of all box coordinate embedding of the sampled object trajectory. FLOPs is negligible because the construction of motion token is just a composition of embedding lookups and token concatenation.

The decoder consists of a multi-head cross-attention(MCA) module and a feed-forward network(FFN). The decoder takes the outputs from the encoder and the motion token as input, generating the final vision-motion representation $\mathbf{f}_{vm}\in\mathbb{R}^{\frac{H_x}{s}\times \frac{W_x}{s}\times C}$ of by computing cross-attention over $\mathbf{f}_{x}^{L}$ and ${\rm Concat}(\mathbf{E}_{motion}, \mathbf{f}_{z}^{L},\mathbf{f}_{x}^{L})$. The decoder is akin to a layer in the encoder, except that the correlation between the template tokens and the search tokens is dropped since we do not need to update the features from the template image in the last layer. The process of the decoder is formulated as:

	\begin{equation}
	\begin{split}
	\mathbf{f}_{m}^{D} & = {\rm Concat}(\mathbf{E}_{motion}, \mathbf{f}_{z}^{L},\mathbf{f}_{x}^{L}) \\
	\mathbf{f}_{vm}' & = \mathbf{f}_{x}^{L}+{\rm MCA}({\rm LN}(\mathbf{f}_{x}^{L}), {\rm LN}(\mathbf{f}_{m}^{D})) \\
	\mathbf{f}_{vm} & = \mathbf{f}_{vm}'+{\rm FFN}({\rm LN}(\mathbf{f}_{vm}')).
	\end{split}
	\end{equation}

$\mathbf{f}_{vm}$ will feed to the head network to generate a classification response map and a bounding box regression map.

{\bf Positional encoding.} Transformer requires a positional encoding to identify the position of the current processing token\cite{vaswani2017attention} because the self-attention module is permutation-invariance. We adopt the \emph{untied positional encoding} \cite{ke2021rethinking} as our positional encoding method. The \emph{untied positional encoding} enhances the expressiveness of the model through untie the positional embeddings from token embeddings with an isolated positional embedding matrix. It also considers the case of special tokens, like the motion token in this paper. We generalize the \emph{untied positional encoding} to multi-dimensions multi-sources data to comply with \emph{concatenated-based fusion} in our tracker. See the appendix for the details.

\subsection{Discussion}

{\bf Why concatenated attention?} To simplify the description, we call the method described above \emph{concatenation-based fusion}. To fuse and process features from multiple sources, it is intuitive to perform self-attention on the feature from each source separately and then compute cross-attention across features from different sources. We call this method \emph{cross-attention-based fusion}. Transformer makes fewer assumptions about the spatial structure of data, which provides great modeling flexibility. In comparison to \emph{cross-attention-based fusion}, \emph{concatenation-based fusion} can save computation cost through operation sharing and reduce model parameters through weight sharing. From the perspective of metric learning, weight sharing is an essential design to ensure the metric between two branches of data is symmetric. Through \emph{concatenation-based fusion}, we implement this property not only in the feature extraction stage but also in the feature fusion stage. In general, \emph{concatenation-based fusion} improves both efficiency and performance. 

{\bf Why not window-based self/cross-attention?} Since we select stage 3 of the Swin-Transformer as the output, the number of tokens involved is significantly reduced, window-based attention cannot save too many FLOPs. Furthermore, considering the extra latency introduced by the window partition and window reverse operations, window-based attention may even be the slower one.

{\bf Why not a query-based decoder?} Derivated from vanilla Transformer decoder, many transformer-based models in vision tasks leverage a learnable query to extract the desired objective features from the encoder, like object queries in \cite{carion2020enddetr}, target query in \cite{stark}. However, in our experiment, a query-based decoder suffers from slow convergence and inferior performance. Most Siamese trackers~\cite{li2018highsiamrpn, xu2020siamfc++, han2021learning} formulate tracking as a foreground-background classification problem, which can better exploit the background information. The vanilla Transformer decoder is a generative model, the generative approaches are considered not suitable for the classification tasks. In another aspect, learning a general target query for any kind of object might cause a bottleneck. In terms of vanilla Transformer encoder-decoder architecture, SwinTrack is an "encoder" only model. Furthermore, quite a little domain knowledge can be easily applied on a classic Siamese tracker to improve the performance, like introducing the smooth movement assumption by using Hanning penalty window on the response map.

{\bf Are other forms of motion token feasible?} Other forms to construct motion token are possible, such as constructing motion token by summing up the past box coordinate embeddings or representing past object trajectories by one token per box. In our early experiments, we find that the proposed motion token is more effective with the best performance. Summing up the past box coordinate embeddings may result in over-parameterization on the coordinate embeddings. While adding temporal motion representation along with visual features to the single-layer decoder in a multi-token form is ineffective, precise temporal modeling may be required in this form.

\subsection{Head and Loss}

{\bf Head.} The head network is split into two branches: classification and bounding box regression. Each of them is a three-layer perceptron. And both of them receives the feature map from the decoder as input to predict the classification response map $r_{cls}\in\mathbb{R}^{(H_x\times W_x)\times 1}$ and bounding box regression map $r_{reg}\in\mathbb{R}^{(H_x\times W_x)\times 4}$, respectively.

{\bf Classification loss.} In classification branch, we employ the \emph{IoU-aware classification score} as the training target and the \emph{varifocal loss}~\cite{zhang2020varifocalnet} as the training loss function. IoU-aware design has been very popular recently, but most works consider IoU prediction as an auxiliary branch to assist classification or bounding box regression~\cite{Ocean_2020_ECCV,bhat2019learning,xu2020siamfc++}. To remove the gap between different prediction branches, \cite{zhang2020varifocalnet} and \cite{li2020generalizedfocal} replace the hard classification target from the ground-truth value, (i.e., 1 for positive samples, 0 for negative samples), to the IoU between the predicted bounding box and the ground-truth one, which is named the \emph{IoU-aware classification score} (IACS). IACS can help the model select a more accurate bounding box prediction candidate from the pool by trying to predict the quality of the bounding box prediction in another branch at the same position. Along with the IACS, the varifocal loss was proposed in \cite{zhang2020varifocalnet} to help the IACS approach outperform other IoU-aware designs. 

The classification loss can be formulated as:
\begin{equation}
    \mathbb{L}_{cls}=\mathbb{L}_{\rm VFL}(p, {\rm IoU}(b, \hat{b})),
\end{equation}
where $p$ denotes the predicted IACS, $b$ denotes the predicted bounding box, and $\hat{b}$ denotes the ground-truth bounding box.

{\bf Regression loss.} For bounding box regression, we employ the generalized IoU loss\cite{GIoU}. The regression loss function can be formulated as:
\begin{equation}
    \mathbb{L}_{reg}=\sum_j \mathbbm{1}_{\{{\rm IoU}(b_j,\hat{b})>0\}}[p\mathbb{L}_{\rm GIoU}(b_j,\hat{b})].
\end{equation}
The GIoU loss is weighted by $p$ to emphasize the high classification score samples. The training signals from the negative samples are ignored.

\section{Experiments}

\subsection{Implementation}

{\bf Model.} We design two variants of SwinTrack with different configurations as follows:
\begin{itemize}
    \item {\bf SwinTrack-T-224}. \\
    Backbone: Swin Transformer-Tiny~\cite{liu2021Swin}, pretrained with ImageNet-1k; \\
    Template size: $[112\times112]$; Search region size: $[224\times224]$; 
    $C=384$; $N=4$;
    \item {\bf SwinTrack-B-384}. \\
    Backbone: Swin Transformer-Base~\cite{liu2021Swin}, pretrained with ImageNet-22k; \\
    Template size: $[192\times192]$; 
    Search region size: $[384\times384]$; 
    $C=512$; $N=8$;
\end{itemize}
where $C$ and $N$ are the channel number of the hidden layers in the first stage of Swin Transformer and the number of encoder blocks in feature fusion, respectively. In all variants, we use the output after the third stage of Swin Transformer for feature extraction. Thus, the backbone stride $s$ is 16.

For motion token, the number of sampled object trajectory $n$ is set to 16, the fixed sampling interval $\Delta$ is set to 15. If the frame rate of the video sequence is available, the sampling interval is adjusted according to the frame rate. Suppose the frame rate is $\mathbbm{f}$, the new sampling interval is getting by $\frac{\Delta}{30}\mathbbm{f}$, 30 fps is the standard frame rate we assumed. $\mathbbm{g}$, which controls the embedding granularity, is set to the same size as the search region feature map, like 14 for SwinTrack-T-224, and 24 for SwinTrack-B-384. For the model for GOT-10k sequences, $n$ is set to 8, $\Delta$ is set to 8, and no frame rate adjustment is applied.

{\bf Training.} We train SwinTrack using the training splits of LaSOT~\cite{fan2019lasot}, TrackingNet~\cite{muller2018trackingnet}, GOT-10k~\cite{Huang2021got10k} (1,000 videos are removed following~\cite{stark} for fair comparison) and COCO 2017~\cite{lin2014microsoftcoco}. In addition, we report the results of SwinTrack-T-224 and SwinTrack-B-384 with GOT-10k training split only to follow the protocol described in~\cite{Huang2021got10k}.

The model is optimized with AdamW~\cite{loshchilov2017decoupledadamw}, with a learning rate of 5e-4, and a weight decay of 1e-4. The learning rate of the backbone is set to 5e-5. We train the network on 8 NVIDIA V100 GPUs for 300 epochs with 131,072 samples per epoch. The learning rate is dropped by a factor of 10 after 210 epochs. A 3-epoch linear warmup is applied to stabilize the training process. DropPath~\cite{larsson2016fractalnet} is applied on the backbone and the encoder with a rate of 0.1. For the models trained for the GOT-10k evaluation protocol, to prevent over-fitting, the training epoch is set to 150, and the learning rate is dropped after 120 epochs. 

For the motion token, the object trajectory for the Siamese training pair is generated with the method described above. The frames that object annotated as absent or out of the video sequence are marked as invalid, the corresponding box coordinates set to $-\infty$. Since the coarse granularity of the coordinate embedding in our setting is already can be seen as an augmentation of historical object trajectory, no additional data augmentation is applied.

{\bf Inference.} We follow the common procedures for Siamese network-based tracking~\cite{bertinetto2016siamfc}. The template image is cropped from the first frame of the video sequence. The target object is in the center of the image with a background area factor of 2. The search region is cropped from the current tracking frame, and the image center is the target center position predicted in the previous frame. The background area factor for the search region is 4.

Our SwinTrack takes the template image and search region as inputs and output classification map $r_{cls}$ and regression map $r_{reg}$. To utilize positional prior in tracking, we apply hanning window penalty on $r_{cls}$, and the final classification map $r_{cls}'$ is obtained via $r_{cls}'=(1-\gamma)\times r_{cls} + \gamma\times h$, where $\gamma$ is the weight parameter and $h$ is the Hanning window with the same size as $r_{cls}$. The target position is determined by the largest value in $r_{cls}'$ and the scale is estimated based on the corresponding regression results in $r_{reg}$.

For the motion token, the predicted confidence score and bounding box are collected on the fly. A confidence threshold $\theta_{conf}$ is applied, if the confidence score given by the classification branch of the head is lower than the threshold, the target object in the current frame is marked as lost by setting the collected bounding box to $-\infty$. $\theta_{conf}$ is set to $0.4$ for LaSOT, the rests are set to $0.3$.

\subsection{Comparisons to State-of-the-arts}

We conduct experiments and compare SwinTrack with SoTA trackers on five benchmarks.

\begin{table}[t]
\caption{Experiments and comparisons on five benchmarks: LaSOT, LaSOT$_{\text{ext}}$, TrackingNet, GOT-10k and TNL2k.}
\label{sota_compare}
\begin{center}
\resizebox{\linewidth}{!}{
\begin{tabular}{r cc cc cc ccc cc}
\hline
\multirow{2}{*}{Tracker} &
  \multicolumn{2}{c}{LaSOT~\cite{fan2019lasot}} &
  \multicolumn{2}{c}{LaSOT$_{\text{ext}}$~\cite{fan2021lasot}} &
  \multicolumn{2}{c}{TrackingNet~\cite{muller2018trackingnet}} &
  \multicolumn{3}{c}{GOT-10k~\cite{Huang2021got10k}} &
  \multicolumn{2}{c}{TNL2k~\cite{wang2021tnl2k}} \\
                                        & SUC  & P    & SUC  & P    & SUC  & P    & AO   & SR$_{0.5}$ & SR$_{0.75}$ & SUC  & P \\ \hline \hline
C-RPN~\cite{Fan_2019_CVPR}               & 45.5 & 44.3 & 27.5 & 32.0 & 66.9 & 61.9 & -    & -          & -           & -    & -  \\ 
SiamPRN++~\cite{li2019evolutionsiamrpnp} & 49.6 & 49.1 & 34.0 & 39.6 & 73.3 & 69.4 & 51.7 & 61.6       & 32.5        & 41.3 & 41.2 \\
Ocean~\cite{Ocean_2020_ECCV}            & 56.0 & 56.6 & -    & -    & -    & -    & 61.1 & 72.1       & 47.3        & 38.4 & 37.7 \\ 
DiMP~\cite{bhat2019learning}            & 56.9 & 56.7 & 39.2 & 45.1 & 74.0 & 68.7 & 61.1 & 71.7       & 49.2        & 44.7 & 43.4 \\
LTMU~\cite{dai2020high}                 & 57.2 & 57.2 & 41.4 & 47.3 & -    & -    & -    & -          & -           & 48.5 & 47.3 \\
SiamR-CNN~\cite{voigtlaender2020siam}   & 64.8 & -    & -    & -    & 81.2 & 80.0 & 64.9 & 72.8       & 59.7        & 52.3 & 52.8 \\ 
STMTrack~\cite{fu2021stmtrack}          & 60.6 & 63.3 & -    & -    & 80.3 & 76.7 & 64.2 & 73.7       & 57.5        & -    & -    \\ 
AutoMatch~\cite{zhang2021learn}         & 58.3 & 59.9 & 37.6 & 43.0 & 76.0 & 72.6 & 65.2 & 76.6       & 54.3        & -    & -    \\
TrDiMP~\cite{wang2021transformer}       & 63.9 & 61.4 & -    & -    & 78.4 & 73.1 & 67.1 & 77.7       & 58.3        & -    & -    \\ 
TransT~\cite{TransT}                    & 64.9 & 69.0 & -    & -    & 81.4 & 80.3 & 67.1 & 76.8       & 60.9        & 51.0 & -    \\ 
STARK~\cite{stark}                      & 67.1 & -    & -    & -    & 82.0 & -    & 68.8 & 78.1       & 64.1        & -    & -    \\ 
KeepTrack~\cite{keeptrack}              & 67.1 & 70.2 & 48.2 & -    & -    & -    & -    & -          & -           & -    & -    \\ 

\hline

SwinTrack-T-224                         & 67.2 & 70.8 & 47.6 & 53.9 & 81.1 & 78.4 & 71.3 & 81.9       & 64.5        & 53.0 & 53.2 \\
SwinTrack-B-384                         & 71.3 & 76.5 & 49.1 & 55.6 & 84.0 & 82.8 & 72.4 & 80.5       & 67.8        & 55.9 & 57.1 \\
\hline
 \end{tabular}
}
\end{center}
\end{table}

{\bf LaSOT.} LaSOT~\cite{fan2019lasot} consists of 280 videos for test. Tab.~\ref{sota_compare} shows the results and comparisons with SoTAs. From Tab.~\ref{sota_compare}, we can observe that SwinTrack-T-224 with light architecture reaches SoTA performance with 0.672 SUC and 0.708 PRE scores, which is competitive compared with other Transformer-based trackers, including STARK-ST101 (0.671 SUC score) and TransT (0.649 SUC), and other trackers using complicated designs such as KeepTrack (0.671 SUC) and SiamR-CNN (0.648 SUC score). With a larger backbone and input size, our strongest variant SwinTrack-B-384 sets a new record with 0.713 SUC score, surpassing START-ST101 and KeepTrack by 4.2 absolute percentage points.

{\bf LaSOT$_{\mathrm{ext}}$.} The recent LaSOT$_{\mathrm{ext}}$~\cite{fan2021lasot} is an extension of LaSOT by adding 150 extra videos. These new sequences are challenging as many similar distractors cause difficulties for tracking. The results of our tracker related to this dataset are an average of three times. KeepTrack uses a complex association technique to handle distractors and achieves a promising 0.482 SUC score as in Tab.~\ref{sota_compare}. Compared with complicated KeepTrack, SwinTrack-T-224 is simple and neat, yet shows comparable performance with 0.476 SUC score. In addition, due to complicated design, KeepTrack runs at less than 20 {\it fps}, while SwinTrack-T-224 runs in 98 {\it fps}, 5$\times$ faster than KeepTrack. When using a larger model, SwinTrack-B-384 shows the best performance with 0.491 SUC score.

{\bf TrackingNet.} We evaluate different trackers on the test set of TrackingNet~\cite{muller2018trackingnet}. From Tab.~\ref{sota_compare}, we observe that our SwinTrack-T-224 achieves a comparable result with 0.811 SUC score. Using a larger model and input size, SwinTrack-B-384 obtains the best performance with 0.840 SUC score, better than STARK-ST101 with 0.820 SUC score and TransT with 0.814 SUC score.

{\bf GOT-10k.} GOT-10k~\cite{Huang2021got10k} offers 180 videos for test and it requires trackers to be trained using GOT-10k train split only. From Tab.~\ref{sota_compare}, we see that SwinTrack-B-384 achieves the best mAO of 0.724, and SwinTrack-T-224 obtains a mAO of 0.713. Both models outperform other Transformer-based counterparts significantly, including START-ST101 (0.688 mAO), TransT (0.671 mAO) and TrDiMP (0.671 mAO).

{\bf TNL2k.} TNL2k~\cite{wang2021tnl2k} is a newly released tracking dataset with 700 videos for test. As reported in Tab.~\ref{sota_compare}, both models surpass the others. SwinTrack-B-384 set a new state-of-the-art with 0.559 SUC score.

{\bf Efficiency comparison.} We report the comparisons of SwinTrack with other Transformer-based trackers in terms of efficiency and complexity. As displayed in Tab.~\ref{efficiency_table}, SwinTrack-T-224 with a small model runs the fastest with a speed of {\it 98 fps}. Especially, compared with STARK-ST101 and STARK-ST50 with 32 {\it fps} and 42 {\it fps}, SwinTrack-T-224 is 3$\times$ and 2$\times$ faster. Despite using a larger model, our SwinTrack-B-384 is still faster than STARK-ST101 and STARK-ST50.

\begin{table}
\caption{Comparison on running speed and $\#$ parameters with other Transformer-based trackers.}
\label{efficiency_table}
\centering
\begin{tabular}{@{}rcccc@{}}
\hline
Tracker & Speed ({\it fps}) & MACs\tablefootnote{Multiply–accumulate operation} (G) & Params (M) \\
\hline \hline
TrDiMP~\cite{wang2021transformer} & 26 & - & - \\
TransT~\cite{TransT} & 50 & - & 23 \\
STARK-ST50~\cite{stark} & 42 & 10.9 & 24 \\
STARK-ST101~\cite{stark} & 32 & 18.5 & 42 \\
\hline
SwinTrack-T-224 & 98 & 6.4 & 23 \\
SwinTrack-B-384 & 45 & 69.7 & 91 \\
\hline
\end{tabular}
\end{table}

\subsection{Ablation Experiment}

\begin{table}
  \caption{Ablation experiments of SwinTrack on four benchmarks. The experiments are conducted on SwinTrack-T-224 without the motion token. \ding{182}: baseline method, \ie, SwinTrack-T-224 without motion token; \ding{183}: replacing Transformer backbone in the baseline method with ResNet-50; \ding{184}: replacing our feature fusion with cross attention-based fusion in the baseline method; \ding{185}: replacing the decoder in baseline with a target query-based; \ding{186}: replacing united positional encoding with absolute sine position encoding in the baseline method; \ding{187}: replacing the IoU-aware classification loss with the plain binary cross entropy loss; \ding{188}: removing the Hanning penalty window in the baseline method inference. }
  \label{tiny_ablation}
  \centering
    \begin{tabular}{ccccccc}
    \hline
          & \tabincell{c}{LaSOT\\SUC (\%)} & \tabincell{c}{LaSOT$_{\mathrm{ext}}$ \\ SUC (\%)} & \tabincell{c}{TrackingNet \\ SUC (\%)} &\tabincell{c}{GOT-10k\tablefootnote{The GOT-10k results in this column are trained with full training datasets.} \\ mAO (\%)} & \tabincell{c}{Speed \\ {\it fps}} & \tabincell{c}{Params \\ M} \\
    \hline\hline
    \ding{182}
    & 66.7  & 46.9  & 80.8  & 70.9  & 98 & 22.7 \\
    \ding{183}
    & 64.2  & 41.8  & 79.5  & 68.2  & 121 & 20.0 \\
    \ding{184}
    & 66.6  & 45.4  & 80.2  & 69.3  & 72 & 34.6 \\
    \ding{185}
    & 66.6  & 43.2  & 79.6  & 69.0  & 91 & 25.3 \\
    \ding{186}
    & 65.7  & 45.0  & 80.0  & 70.0  & 103 & 21.6 \\
    \ding{187}
    & 66.2  & 46.7  & 79.4  & 68.2  & 98 & 22.7 \\
    \ding{188}
    & 65.7  & 46.0  & 80.0  & 69.6  & 98 & 22.7 \\
    \hline
    \end{tabular}%
\end{table}%

{\bf Comparison with ResNet backbone.} We compare the Swin-Transformer backbone with popular ResNet-50~\cite{he2016deep}. As shown in Tab.~\ref{tiny_ablation} (\ding{182} {\it vs.} \ding{183}). The Swin Transformer backbone significantly boosts the performance by 2.5\% SUC score in LaSOT, 5.1\% SUC score in LaSOT$_\textrm{ext}$. The result shows that the strong appearance modeling capability provided by the Swin Transformer plays a crucial role.

{\bf Feature fusion.} As displayed in Tab.~\ref{tiny_ablation} (\ding{182} {\it vs.} \ding{184}), compared with the \emph{concatenation-based fusion}, the \emph{cross attention-based fusion} runs at a slower speed, occupies much more memory, and also has an inferior performance on all datasets. Slower speed can be due to the latency brought by the extra operations. The parameter-sharing strategy not only just reduces the number of parameters but also benefits metric learning.

{\bf Comparison with the query-based decoder.} Queries is commonly adopted in the decoder of Transformer network in vision tasks, e.g. object query~\cite{carion2020enddetr} and target query~\cite{stark}. Nevertheless, our empirical results in Tab.~\ref{tiny_ablation} (\ding{182} {\it vs.} \ding{185}) show that a target query-based decoder degrades the tracking performance on all benchmarks, even with $2 \times$ training pairs. As discussed, one possible reason is the generative model is not suitable for classification. Besides, learning a general target query for any kind of object may also be difficult.

{\bf Position encoding.} We compare the united positional encoding used in SwinTrack and the original absolute position encoding in Transformer~\cite{vaswani2017attention}. Notice, We make a little modification to the original absolute position encoding: Except for the 2D embedding, the index of token source (e.g. 1 for the tokens from the template patch, 2 for the tokens from the search region patch) is also embedded. As shown in Tab.~\ref{tiny_ablation} (\ding{182} {\it vs.} \ding{186}), our method with united positional encoding obtains improvements with 0.8-1.9 absolute percentage points on the benchmarks with negligible loss in speed (98 {\it vs.} 103).

{\bf Loss function.} From Tab.~\ref{tiny_ablation} (\ding{182} {\it vs.} \ding{187}), we observe that the model trained with varifocal loss significantly outperforms the one with binary cross entropy (BCE) loss without loss of efficiency. This result indicates that the varifocal loss can assist the classification branch of the head to generate an IoU-aware response map, and thus help the tracker to improve the tracking performance.

{\bf Post processing.} One may wonder with highly discriminative Transformer architecture and IoU-aware classification loss does the hanning penalty window is still functional, which introduces a strong smooth movement assumption. In the experiments, we remove the hanning penalty window in post-processing, as shown in Tab.~\ref{tiny_ablation} (\ding{182} {\it vs.} \ding{188}), the performance is dropped by 1.0 SUC for LaSOT, 1.3 AO for GOT-10k in absolute percentage, and less than 1\% in the SUC metric of other datasets. This suggests that the strong smooth movement assumption is still applicable for our tracker. But compared with the former Transformer-based tracker \cite{TransT}, the performance gap between with and without penalty window post-processing is narrowing.

\begin{table}
  \caption{Ablation experiments on our proposed motion token on the tracking performance on four benchmarks. The experiments are conducted on SwinTrack-T-224. \ding{182}: SwinTrack-T-224; \ding{183}: SwinTrack-B-384; \ding{184}: SwinTrack-T-224 without motion token; \ding{185}: SwinTrack-B-384 without motion token; \ding{186}: replacing the motion token in SwinTrack-T-224 with a learnable embedding token.
  }
  \label{motion_token_ablation}
  \centering
    \begin{tabular}{@{}cccccc@{}}
    \hline
          & \tabincell{c}{LaSOT\\SUC (\%)} & \tabincell{c}{LaSOT$_{\textrm{ext}}$ \\ SUC (\%)} & \tabincell{c}{TrackingNet \\ SUC (\%)} &\tabincell{c}{GOT-10k \\ mAO (\%)} & \tabincell{c}{Speed \\ {\it fps}} \\
    \hline\hline
    \ding{182}
    & 67.2  & 47.6  & 81.1  & 71.3  & 96 \\
    \ding{183}
    & 71.3  & 49.1  & 84.0  & 72.4  & 45   \\
    \hline
    \ding{184}
    & 66.7  & 47.0  & 80.8  & 70.0  & 98 \\
    \ding{185}
    & 70.2  & 48.5  & 84.0  & 70.7  & 45 \\
    \hline
    \ding{186}
    & 66.3  & 45.2  & 81.2  & 70.0  & 96 \\
    \hline
    \end{tabular}%
\end{table}%

{\bf Effectiveness of motion token.} We study the effectiveness of the motion token by conducting comparison experiments. As shown in Tab.~\ref{motion_token_ablation} (\ding{182} vs. \ding{184} and \ding{183} vs. \ding{185}), the models with motion token outperforms the models without motion token on all datasets, especially on LaSOT$_{\textrm{ext}}$ and GOT-10k. The results indicate that the motion token can assist the tracker to handle hard similar distractors in LaSOT$_{\textrm{ext}}$ and stabilize the short-term tracking like the sequences in GOT-10k test set. We also study whether the effectiveness of the motion token is simply from the extra embedding vector. We set up an experiment as in Tab.~\ref{motion_token_ablation} (\ding{186}), which replaces the motion token with a learnable embedding token. The result shows that the extra embedding vector has negative impacts indicating the effectiveness of the embedding of object trajectory.

\section{Conclusion}

In this work, we present SwinTrack, a simple and strong baseline for Transformer tracking. In SwinTrack, both representation learning and feature fusion are implemented with the attention mechanism. Extensive experiments demonstrate the effectiveness of such architecture. Besides, we propose the motion token to enhance the robustness of the tracker by providing the historical object trajectory, showing the flexibility of the Transformer model in architectural design. With the power of sequence-to-sequence model architecture, a context-rich tracker is possible, and more contextual cues can be incorporated. Finally, We hope this work can inspire and facilitate future research.

\begin{ack}
This work is supported by Peng Cheng Laboratory Research Project No. PCL2021A07. Heng Fan and his employer receives no financial support for the
research, authorship, and/or publication of this article.
\end{ack}

{
\small
\bibliographystyle{elsarticle-harv}
\bibliography{final}

\begin{thebibliography}{52}
\expandafter\ifx\csname natexlab\endcsname\relax\def\natexlab#1{#1}\fi
\providecommand{\url}[1]{\texttt{#1}}
\providecommand{\href}[2]{#2}
\providecommand{\path}[1]{#1}
\providecommand{\DOIprefix}{doi:}
\providecommand{\ArXivprefix}{arXiv:}
\providecommand{\URLprefix}{URL: }
\providecommand{\Pubmedprefix}{pmid:}
\providecommand{\doi}[1]{\href{http://dx.doi.org/#1}{\path{#1}}}
\providecommand{\Pubmed}[1]{\href{pmid:#1}{\path{#1}}}
\providecommand{\bibinfo}[2]{#2}
\ifx\xfnm\relax \def\xfnm[#1]{\unskip,\space#1}\fi
\bibitem[{Bertinetto et~al.(2016)Bertinetto, Valmadre, Henriques, Vedaldi and
  Torr}]{bertinetto2016siamfc}
\bibinfo{author}{Bertinetto, L.}, \bibinfo{author}{Valmadre, J.},
  \bibinfo{author}{Henriques, J.F.}, \bibinfo{author}{Vedaldi, A.},
  \bibinfo{author}{Torr, P.H.}, \bibinfo{year}{2016}.
\newblock \bibinfo{title}{Fully-convolutional siamese networks for object
  tracking}, in: \bibinfo{booktitle}{ECCVW}.
\bibitem[{Bhat et~al.(2019)Bhat, Danelljan, Gool and
  Timofte}]{bhat2019learning}
\bibinfo{author}{Bhat, G.}, \bibinfo{author}{Danelljan, M.},
  \bibinfo{author}{Gool, L.V.}, \bibinfo{author}{Timofte, R.},
  \bibinfo{year}{2019}.
\newblock \bibinfo{title}{Learning discriminative model prediction for
  tracking}, in: \bibinfo{booktitle}{ICCV}.
\bibitem[{Carion et~al.(2020)Carion, Massa, Synnaeve, Usunier, Kirillov and
  Zagoruyko}]{carion2020enddetr}
\bibinfo{author}{Carion, N.}, \bibinfo{author}{Massa, F.},
  \bibinfo{author}{Synnaeve, G.}, \bibinfo{author}{Usunier, N.},
  \bibinfo{author}{Kirillov, A.}, \bibinfo{author}{Zagoruyko, S.},
  \bibinfo{year}{2020}.
\newblock \bibinfo{title}{End-to-end object detection with transformers}, in:
  \bibinfo{booktitle}{ECCV}.
\bibitem[{Chen et~al.(2021a)Chen, Fan and Panda}]{chen2021crossvit}
\bibinfo{author}{Chen, C.F.R.}, \bibinfo{author}{Fan, Q.},
  \bibinfo{author}{Panda, R.}, \bibinfo{year}{2021}a.
\newblock \bibinfo{title}{Crossvit: Cross-attention multi-scale vision
  transformer for image classification}, in: \bibinfo{booktitle}{ICCV}.
\bibitem[{Chen et~al.(2021b)Chen, Yan, Zhu, Wang, Yang and Lu}]{TransT}
\bibinfo{author}{Chen, X.}, \bibinfo{author}{Yan, B.}, \bibinfo{author}{Zhu,
  J.}, \bibinfo{author}{Wang, D.}, \bibinfo{author}{Yang, X.},
  \bibinfo{author}{Lu, H.}, \bibinfo{year}{2021}b.
\newblock \bibinfo{title}{Transformer tracking}, in: \bibinfo{booktitle}{CVPR}.
\bibitem[{Cui et~al.(2022)Cui, Jiang, Wang and Wu}]{cui2022mixformer}
\bibinfo{author}{Cui, Y.}, \bibinfo{author}{Jiang, C.}, \bibinfo{author}{Wang,
  L.}, \bibinfo{author}{Wu, G.}, \bibinfo{year}{2022}.
\newblock \bibinfo{title}{Mixformer: End-to-end tracking with iterative mixed
  attention}, in: \bibinfo{booktitle}{CVPR}.
\bibitem[{Dai et~al.(2020)Dai, Zhang, Wang, Li, Lu and Yang}]{dai2020high}
\bibinfo{author}{Dai, K.}, \bibinfo{author}{Zhang, Y.}, \bibinfo{author}{Wang,
  D.}, \bibinfo{author}{Li, J.}, \bibinfo{author}{Lu, H.},
  \bibinfo{author}{Yang, X.}, \bibinfo{year}{2020}.
\newblock \bibinfo{title}{High-performance long-term tracking with
  meta-updater}, in: \bibinfo{booktitle}{CVPR}.
\bibitem[{Danelljan et~al.(2019)Danelljan, Bhat, Khan and
  Felsberg}]{danelljan2019atom}
\bibinfo{author}{Danelljan, M.}, \bibinfo{author}{Bhat, G.},
  \bibinfo{author}{Khan, F.S.}, \bibinfo{author}{Felsberg, M.},
  \bibinfo{year}{2019}.
\newblock \bibinfo{title}{Atom: Accurate tracking by overlap maximization}, in:
  \bibinfo{booktitle}{CVPR}.
\bibitem[{Dosovitskiy et~al.(2021)Dosovitskiy, Beyer, Kolesnikov, Weissenborn,
  Zhai, Unterthiner, Dehghani, Minderer, Heigold, Gelly
  et~al.}]{dosovitskiy2020imagevit}
\bibinfo{author}{Dosovitskiy, A.}, \bibinfo{author}{Beyer, L.},
  \bibinfo{author}{Kolesnikov, A.}, \bibinfo{author}{Weissenborn, D.},
  \bibinfo{author}{Zhai, X.}, \bibinfo{author}{Unterthiner, T.},
  \bibinfo{author}{Dehghani, M.}, \bibinfo{author}{Minderer, M.},
  \bibinfo{author}{Heigold, G.}, \bibinfo{author}{Gelly, S.}, et~al.,
  \bibinfo{year}{2021}.
\newblock \bibinfo{title}{An image is worth 16x16 words: Transformers for image
  recognition at scale}, in: \bibinfo{booktitle}{ICLR}.
\bibitem[{Fan et~al.(2021)Fan, Bai, Lin, Yang, Chu, Deng, Yu, Huang, Liu, Xu
  et~al.}]{fan2021lasot}
\bibinfo{author}{Fan, H.}, \bibinfo{author}{Bai, H.}, \bibinfo{author}{Lin,
  L.}, \bibinfo{author}{Yang, F.}, \bibinfo{author}{Chu, P.},
  \bibinfo{author}{Deng, G.}, \bibinfo{author}{Yu, S.}, \bibinfo{author}{Huang,
  M.}, \bibinfo{author}{Liu, J.}, \bibinfo{author}{Xu, Y.}, et~al.,
  \bibinfo{year}{2021}.
\newblock \bibinfo{title}{Lasot: A high-quality large-scale single object
  tracking benchmark}.
\newblock \bibinfo{journal}{International Journal of Computer Vision}
  \bibinfo{volume}{129}, \bibinfo{pages}{439--461}.
\bibitem[{Fan et~al.(2019)Fan, Lin, Yang, Chu, Deng, Yu, Bai, Xu, Liao and
  Ling}]{fan2019lasot}
\bibinfo{author}{Fan, H.}, \bibinfo{author}{Lin, L.}, \bibinfo{author}{Yang,
  F.}, \bibinfo{author}{Chu, P.}, \bibinfo{author}{Deng, G.},
  \bibinfo{author}{Yu, S.}, \bibinfo{author}{Bai, H.}, \bibinfo{author}{Xu,
  Y.}, \bibinfo{author}{Liao, C.}, \bibinfo{author}{Ling, H.},
  \bibinfo{year}{2019}.
\newblock \bibinfo{title}{Lasot: A high-quality benchmark for large-scale
  single object tracking}, in: \bibinfo{booktitle}{CVPR}.
\bibitem[{Fan and Ling(2019)}]{Fan_2019_CVPR}
\bibinfo{author}{Fan, H.}, \bibinfo{author}{Ling, H.}, \bibinfo{year}{2019}.
\newblock \bibinfo{title}{Siamese cascaded region proposal networks for
  real-time visual tracking}, in: \bibinfo{booktitle}{CVPR}.
\bibitem[{Fan and Ling(2021)}]{fan2020cract}
\bibinfo{author}{Fan, H.}, \bibinfo{author}{Ling, H.}, \bibinfo{year}{2021}.
\newblock \bibinfo{title}{Cract: Cascaded regression-align-classification for
  robust visual tracking}, in: \bibinfo{booktitle}{IROS}.
\bibitem[{Fu et~al.(2021)Fu, Liu, Fu and Wang}]{fu2021stmtrack}
\bibinfo{author}{Fu, Z.}, \bibinfo{author}{Liu, Q.}, \bibinfo{author}{Fu, Z.},
  \bibinfo{author}{Wang, Y.}, \bibinfo{year}{2021}.
\newblock \bibinfo{title}{Stmtrack: Template-free visual tracking with
  space-time memory networks}, in: \bibinfo{booktitle}{CVPR}.
\bibitem[{Gao et~al.(2022)Gao, Zhou, Ma, Wang and Yuan}]{gao2022aiatrack}
\bibinfo{author}{Gao, S.}, \bibinfo{author}{Zhou, C.}, \bibinfo{author}{Ma,
  C.}, \bibinfo{author}{Wang, X.}, \bibinfo{author}{Yuan, J.},
  \bibinfo{year}{2022}.
\newblock \bibinfo{title}{Aiatrack: Attention in attention for transformer
  visual tracking} .
\bibitem[{Han et~al.(2021)Han, Dong, Khan, Shao and Shen}]{han2021learning}
\bibinfo{author}{Han, W.}, \bibinfo{author}{Dong, X.}, \bibinfo{author}{Khan,
  F.S.}, \bibinfo{author}{Shao, L.}, \bibinfo{author}{Shen, J.},
  \bibinfo{year}{2021}.
\newblock \bibinfo{title}{Learning to fuse asymmetric feature maps in siamese
  trackers}, in: \bibinfo{booktitle}{CVPR}.
\bibitem[{He et~al.(2022)He, Chen, Xie, Li, Doll{\'a}r and
  Girshick}]{he2022masked}
\bibinfo{author}{He, K.}, \bibinfo{author}{Chen, X.}, \bibinfo{author}{Xie,
  S.}, \bibinfo{author}{Li, Y.}, \bibinfo{author}{Doll{\'a}r, P.},
  \bibinfo{author}{Girshick, R.}, \bibinfo{year}{2022}.
\newblock \bibinfo{title}{Masked autoencoders are scalable vision learners},
  in: \bibinfo{booktitle}{CVPR}.
\bibitem[{He et~al.(2016)He, Zhang, Ren and Sun}]{he2016deep}
\bibinfo{author}{He, K.}, \bibinfo{author}{Zhang, X.}, \bibinfo{author}{Ren,
  S.}, \bibinfo{author}{Sun, J.}, \bibinfo{year}{2016}.
\newblock \bibinfo{title}{Deep residual learning for image recognition}, in:
  \bibinfo{booktitle}{CVPR}.
\bibitem[{Huang et~al.(2019)Huang, Zhao and Huang}]{Huang2021got10k}
\bibinfo{author}{Huang, L.}, \bibinfo{author}{Zhao, X.},
  \bibinfo{author}{Huang, K.}, \bibinfo{year}{2019}.
\newblock \bibinfo{title}{Got-10k: A large high-diversity benchmark for generic
  object tracking in the wild}.
\newblock \bibinfo{journal}{IEEE Transactions on Pattern Analysis and Machine
  Intelligence} \bibinfo{volume}{43}, \bibinfo{pages}{1562--1577}.
\bibitem[{Ke et~al.(2021)Ke, He and Liu}]{ke2021rethinking}
\bibinfo{author}{Ke, G.}, \bibinfo{author}{He, D.}, \bibinfo{author}{Liu,
  T.Y.}, \bibinfo{year}{2021}.
\newblock \bibinfo{title}{Rethinking positional encoding in language
  pre-training}, in: \bibinfo{booktitle}{ICLR}.
\bibitem[{Kristan et~al.(2016)Kristan, Matas, Leonardis, Vojir, Pflugfelder,
  Fernandez, Nebehay, Porikli and \v{C}ehovin}]{VOT_TPAMI}
\bibinfo{author}{Kristan, M.}, \bibinfo{author}{Matas, J.},
  \bibinfo{author}{Leonardis, A.}, \bibinfo{author}{Vojir, T.},
  \bibinfo{author}{Pflugfelder, R.}, \bibinfo{author}{Fernandez, G.},
  \bibinfo{author}{Nebehay, G.}, \bibinfo{author}{Porikli, F.},
  \bibinfo{author}{\v{C}ehovin, L.}, \bibinfo{year}{2016}.
\newblock \bibinfo{title}{A novel performance evaluation methodology for
  single-target trackers}.
\newblock \bibinfo{journal}{IEEE Transactions on Pattern Analysis and Machine
  Intelligence} \bibinfo{volume}{38}, \bibinfo{pages}{2137--2155}.
\bibitem[{Krizhevsky et~al.(2012)Krizhevsky, Sutskever and
  Hinton}]{krizhevsky2012imagenet}
\bibinfo{author}{Krizhevsky, A.}, \bibinfo{author}{Sutskever, I.},
  \bibinfo{author}{Hinton, G.E.}, \bibinfo{year}{2012}.
\newblock \bibinfo{title}{Imagenet classification with deep convolutional
  neural networks}.
\newblock \bibinfo{journal}{NIPS} .
\bibitem[{Larsson et~al.(2016)Larsson, Maire and
  Shakhnarovich}]{larsson2016fractalnet}
\bibinfo{author}{Larsson, G.}, \bibinfo{author}{Maire, M.},
  \bibinfo{author}{Shakhnarovich, G.}, \bibinfo{year}{2016}.
\newblock \bibinfo{title}{Fractalnet: Ultra-deep neural networks without
  residuals}, in: \bibinfo{booktitle}{ICLR}.
\bibitem[{Li et~al.(2019)Li, Wu, Wang, Zhang, Xing and
  Yan}]{li2019evolutionsiamrpnp}
\bibinfo{author}{Li, B.}, \bibinfo{author}{Wu, W.}, \bibinfo{author}{Wang, Q.},
  \bibinfo{author}{Zhang, F.}, \bibinfo{author}{Xing, J.},
  \bibinfo{author}{Yan, J.S.}, \bibinfo{year}{2019}.
\newblock \bibinfo{title}{Evolution of siamese visual tracking with very deep
  networks}, in: \bibinfo{booktitle}{CVPR}.
\bibitem[{Li et~al.(2018)Li, Yan, Wu, Zhu and Hu}]{li2018highsiamrpn}
\bibinfo{author}{Li, B.}, \bibinfo{author}{Yan, J.}, \bibinfo{author}{Wu, W.},
  \bibinfo{author}{Zhu, Z.}, \bibinfo{author}{Hu, X.}, \bibinfo{year}{2018}.
\newblock \bibinfo{title}{High performance visual tracking with siamese region
  proposal network}, in: \bibinfo{booktitle}{CVPR}.
\bibitem[{Li et~al.(2020)Li, Wang, Wu, Chen, Hu, Li, Tang and
  Yang}]{li2020generalizedfocal}
\bibinfo{author}{Li, X.}, \bibinfo{author}{Wang, W.}, \bibinfo{author}{Wu, L.},
  \bibinfo{author}{Chen, S.}, \bibinfo{author}{Hu, X.}, \bibinfo{author}{Li,
  J.}, \bibinfo{author}{Tang, J.}, \bibinfo{author}{Yang, J.},
  \bibinfo{year}{2020}.
\newblock \bibinfo{title}{Generalized focal loss: Learning qualified and
  distributed bounding boxes for dense object detection}, in:
  \bibinfo{booktitle}{NeurIPS}.
\bibitem[{Lin et~al.(2014)Lin, Maire, Belongie, Hays, Perona, Ramanan,
  Doll{\'a}r and Zitnick}]{lin2014microsoftcoco}
\bibinfo{author}{Lin, T.Y.}, \bibinfo{author}{Maire, M.},
  \bibinfo{author}{Belongie, S.}, \bibinfo{author}{Hays, J.},
  \bibinfo{author}{Perona, P.}, \bibinfo{author}{Ramanan, D.},
  \bibinfo{author}{Doll{\'a}r, P.}, \bibinfo{author}{Zitnick, C.L.},
  \bibinfo{year}{2014}.
\newblock \bibinfo{title}{Microsoft coco: Common objects in context}, in:
  \bibinfo{booktitle}{ECCV}.
\bibitem[{Liu et~al.(2021)Liu, Lin, Cao, Hu, Wei, Zhang, Lin and
  Guo}]{liu2021Swin}
\bibinfo{author}{Liu, Z.}, \bibinfo{author}{Lin, Y.}, \bibinfo{author}{Cao,
  Y.}, \bibinfo{author}{Hu, H.}, \bibinfo{author}{Wei, Y.},
  \bibinfo{author}{Zhang, Z.}, \bibinfo{author}{Lin, S.}, \bibinfo{author}{Guo,
  B.}, \bibinfo{year}{2021}.
\newblock \bibinfo{title}{Swin transformer: Hierarchical vision transformer
  using shifted windows}.
\newblock \bibinfo{journal}{ICCV} .
\bibitem[{Loshchilov and Hutter(2019)}]{loshchilov2017decoupledadamw}
\bibinfo{author}{Loshchilov, I.}, \bibinfo{author}{Hutter, F.},
  \bibinfo{year}{2019}.
\newblock \bibinfo{title}{Decoupled weight decay regularization}, in:
  \bibinfo{booktitle}{ICLR}.
\bibitem[{Mayer et~al.(2022)Mayer, Danelljan, Bhat, Paul, Paudel, Yu and
  Van~Gool}]{mayer2022transforming}
\bibinfo{author}{Mayer, C.}, \bibinfo{author}{Danelljan, M.},
  \bibinfo{author}{Bhat, G.}, \bibinfo{author}{Paul, M.},
  \bibinfo{author}{Paudel, D.P.}, \bibinfo{author}{Yu, F.},
  \bibinfo{author}{Van~Gool, L.}, \bibinfo{year}{2022}.
\newblock \bibinfo{title}{Transforming model prediction for tracking}, in:
  \bibinfo{booktitle}{CVPR}.
\bibitem[{Mayer et~al.(2021)Mayer, Danelljan, Paudel and Van~Gool}]{keeptrack}
\bibinfo{author}{Mayer, C.}, \bibinfo{author}{Danelljan, M.},
  \bibinfo{author}{Paudel, D.P.}, \bibinfo{author}{Van~Gool, L.},
  \bibinfo{year}{2021}.
\newblock \bibinfo{title}{Learning target candidate association to keep track
  of what not to track}, in: \bibinfo{booktitle}{ICCV}.
\bibitem[{Mueller et~al.(2016)Mueller, Smith and Ghanem}]{mueller2016benchmark}
\bibinfo{author}{Mueller, M.}, \bibinfo{author}{Smith, N.},
  \bibinfo{author}{Ghanem, B.}, \bibinfo{year}{2016}.
\newblock \bibinfo{title}{A benchmark and simulator for uav tracking}, in:
  \bibinfo{booktitle}{ECCV}.
\bibitem[{Muller et~al.(2018)Muller, Bibi, Giancola, Alsubaihi and
  Ghanem}]{muller2018trackingnet}
\bibinfo{author}{Muller, M.}, \bibinfo{author}{Bibi, A.},
  \bibinfo{author}{Giancola, S.}, \bibinfo{author}{Alsubaihi, S.},
  \bibinfo{author}{Ghanem, B.}, \bibinfo{year}{2018}.
\newblock \bibinfo{title}{Trackingnet: A large-scale dataset and benchmark for
  object tracking in the wild}, in: \bibinfo{booktitle}{ECCV}.
\bibitem[{Ren et~al.(2015)Ren, He, Girshick and Sun}]{ren2015faster}
\bibinfo{author}{Ren, S.}, \bibinfo{author}{He, K.}, \bibinfo{author}{Girshick,
  R.}, \bibinfo{author}{Sun, J.}, \bibinfo{year}{2015}.
\newblock \bibinfo{title}{Faster r-cnn: Towards real-time object detection with
  region proposal networks}, in: \bibinfo{booktitle}{NIPS}.
\bibitem[{Rezatofighi et~al.(2019)Rezatofighi, Tsoi, Gwak, Sadeghian, Reid and
  Savarese}]{GIoU}
\bibinfo{author}{Rezatofighi, H.}, \bibinfo{author}{Tsoi, N.},
  \bibinfo{author}{Gwak, J.}, \bibinfo{author}{Sadeghian, A.},
  \bibinfo{author}{Reid, I.}, \bibinfo{author}{Savarese, S.},
  \bibinfo{year}{2019}.
\newblock \bibinfo{title}{Generalized intersection over union} .
\bibitem[{Shaw et~al.(2018)Shaw, Uszkoreit and Vaswani}]{shaw2018relpos}
\bibinfo{author}{Shaw, P.}, \bibinfo{author}{Uszkoreit, J.},
  \bibinfo{author}{Vaswani, A.}, \bibinfo{year}{2018}.
\newblock \bibinfo{title}{Self-attention with relative position
  representations}.
\newblock \bibinfo{journal}{arXiv} .
\bibitem[{Touvron et~al.(2021)Touvron, Cord, Douze, Massa, Sablayrolles and
  J{\'e}gou}]{touvron2021training}
\bibinfo{author}{Touvron, H.}, \bibinfo{author}{Cord, M.},
  \bibinfo{author}{Douze, M.}, \bibinfo{author}{Massa, F.},
  \bibinfo{author}{Sablayrolles, A.}, \bibinfo{author}{J{\'e}gou, H.},
  \bibinfo{year}{2021}.
\newblock \bibinfo{title}{Training data-efficient image transformers \&
  distillation through attention}, in: \bibinfo{booktitle}{ICML}.
\bibitem[{Vaswani et~al.(2017)Vaswani, Shazeer, Parmar, Uszkoreit, Jones,
  Gomez, Kaiser and Polosukhin}]{vaswani2017attention}
\bibinfo{author}{Vaswani, A.}, \bibinfo{author}{Shazeer, N.},
  \bibinfo{author}{Parmar, N.}, \bibinfo{author}{Uszkoreit, J.},
  \bibinfo{author}{Jones, L.}, \bibinfo{author}{Gomez, A.N.},
  \bibinfo{author}{Kaiser, {\L}.}, \bibinfo{author}{Polosukhin, I.},
  \bibinfo{year}{2017}.
\newblock \bibinfo{title}{Attention is all you need}, in:
  \bibinfo{booktitle}{NeurIPS}.
\bibitem[{Voigtlaender et~al.(2020)Voigtlaender, Luiten, Torr and
  Leibe}]{voigtlaender2020siam}
\bibinfo{author}{Voigtlaender, P.}, \bibinfo{author}{Luiten, J.},
  \bibinfo{author}{Torr, P.H.}, \bibinfo{author}{Leibe, B.},
  \bibinfo{year}{2020}.
\newblock \bibinfo{title}{Siam r-cnn: Visual tracking by re-detection}, in:
  \bibinfo{booktitle}{CVPR}.
\bibitem[{Wang et~al.(2021a)Wang, Zhou, Wang and Li}]{wang2021transformer}
\bibinfo{author}{Wang, N.}, \bibinfo{author}{Zhou, W.}, \bibinfo{author}{Wang,
  J.}, \bibinfo{author}{Li, H.}, \bibinfo{year}{2021}a.
\newblock \bibinfo{title}{Transformer meets tracker: Exploiting temporal
  context for robust visual tracking}, in: \bibinfo{booktitle}{CVPR}.
\bibitem[{Wang et~al.(2021b)Wang, Xie, Li, Fan, Song, Liang, Lu, Luo and
  Shao}]{wang2021pyramid}
\bibinfo{author}{Wang, W.}, \bibinfo{author}{Xie, E.}, \bibinfo{author}{Li,
  X.}, \bibinfo{author}{Fan, D.P.}, \bibinfo{author}{Song, K.},
  \bibinfo{author}{Liang, D.}, \bibinfo{author}{Lu, T.}, \bibinfo{author}{Luo,
  P.}, \bibinfo{author}{Shao, L.}, \bibinfo{year}{2021}b.
\newblock \bibinfo{title}{Pyramid vision transformer: A versatile backbone for
  dense prediction without convolutions}, in: \bibinfo{booktitle}{ICCV}.
\bibitem[{Wang et~al.(2021c)Wang, Shu, Zhang, Jiang, Wang, Tian and
  Wu}]{wang2021tnl2k}
\bibinfo{author}{Wang, X.}, \bibinfo{author}{Shu, X.}, \bibinfo{author}{Zhang,
  Z.}, \bibinfo{author}{Jiang, B.}, \bibinfo{author}{Wang, Y.},
  \bibinfo{author}{Tian, Y.}, \bibinfo{author}{Wu, F.}, \bibinfo{year}{2021}c.
\newblock \bibinfo{title}{Towards more flexible and accurate object tracking
  with natural language: Algorithms and benchmark}, in:
  \bibinfo{booktitle}{CVPR}.
\bibitem[{Xie et~al.(2022)Xie, Wang, Wang, Cao, Yang and
  Zeng}]{xie2022correlation}
\bibinfo{author}{Xie, F.}, \bibinfo{author}{Wang, C.}, \bibinfo{author}{Wang,
  G.}, \bibinfo{author}{Cao, Y.}, \bibinfo{author}{Yang, W.},
  \bibinfo{author}{Zeng, W.}, \bibinfo{year}{2022}.
\newblock \bibinfo{title}{Correlation-aware deep tracking}, in:
  \bibinfo{booktitle}{CVPR}.
\bibitem[{Xu et~al.(2020)Xu, Wang, Li, Yuan and Yu}]{xu2020siamfc++}
\bibinfo{author}{Xu, Y.}, \bibinfo{author}{Wang, Z.}, \bibinfo{author}{Li, Z.},
  \bibinfo{author}{Yuan, Y.}, \bibinfo{author}{Yu, G.}, \bibinfo{year}{2020}.
\newblock \bibinfo{title}{Siamfc++: Towards robust and accurate visual tracking
  with target estimation guidelines}, in: \bibinfo{booktitle}{AAAI}.
\bibitem[{Yan et~al.(2022)Yan, Jiang, Sun, Wang, Yuan, Luo and Lu}]{unicorn}
\bibinfo{author}{Yan, B.}, \bibinfo{author}{Jiang, Y.}, \bibinfo{author}{Sun,
  P.}, \bibinfo{author}{Wang, D.}, \bibinfo{author}{Yuan, Z.},
  \bibinfo{author}{Luo, P.}, \bibinfo{author}{Lu, H.}, \bibinfo{year}{2022}.
\newblock \bibinfo{title}{Towards grand unification of object tracking}, in:
  \bibinfo{booktitle}{ECCV}.
\bibitem[{Yan et~al.(2021)Yan, Peng, Fu, Wang and Lu}]{stark}
\bibinfo{author}{Yan, B.}, \bibinfo{author}{Peng, H.}, \bibinfo{author}{Fu,
  J.}, \bibinfo{author}{Wang, D.}, \bibinfo{author}{Lu, H.},
  \bibinfo{year}{2021}.
\newblock \bibinfo{title}{Learning spatio-temporal transformer for visual
  tracking}, in: \bibinfo{booktitle}{ICCV}.
\bibitem[{Ye et~al.(2022)Ye, Chang, Ma, Shan and Chen}]{ye2022ostrack}
\bibinfo{author}{Ye, B.}, \bibinfo{author}{Chang, H.}, \bibinfo{author}{Ma,
  B.}, \bibinfo{author}{Shan, S.}, \bibinfo{author}{Chen, X.},
  \bibinfo{year}{2022}.
\newblock \bibinfo{title}{Joint feature learning and relation modeling for
  tracking: A one-stream framework} .
\bibitem[{Yu et~al.(2020)Yu, Xiong, Huang and Scott}]{yu2020deformable}
\bibinfo{author}{Yu, Y.}, \bibinfo{author}{Xiong, Y.}, \bibinfo{author}{Huang,
  W.}, \bibinfo{author}{Scott, M.R.}, \bibinfo{year}{2020}.
\newblock \bibinfo{title}{Deformable siamese attention networks for visual
  object tracking}, in: \bibinfo{booktitle}{CVPR}.
\bibitem[{Yuan et~al.(2021)Yuan, Chen, Wang, Yu, Shi, Jiang, Tay, Feng and
  Yan}]{yuan2021tokens}
\bibinfo{author}{Yuan, L.}, \bibinfo{author}{Chen, Y.}, \bibinfo{author}{Wang,
  T.}, \bibinfo{author}{Yu, W.}, \bibinfo{author}{Shi, Y.},
  \bibinfo{author}{Jiang, Z.H.}, \bibinfo{author}{Tay, F.E.},
  \bibinfo{author}{Feng, J.}, \bibinfo{author}{Yan, S.}, \bibinfo{year}{2021}.
\newblock \bibinfo{title}{Tokens-to-token vit: Training vision transformers
  from scratch on imagenet}, in: \bibinfo{booktitle}{ICCV}.
\bibitem[{Zhang et~al.(2021a)Zhang, Wang, Dayoub and
  S{\"u}nderhauf}]{zhang2020varifocalnet}
\bibinfo{author}{Zhang, H.}, \bibinfo{author}{Wang, Y.},
  \bibinfo{author}{Dayoub, F.}, \bibinfo{author}{S{\"u}nderhauf, N.},
  \bibinfo{year}{2021}a.
\newblock \bibinfo{title}{Varifocalnet: An iou-aware dense object detector},
  in: \bibinfo{booktitle}{CVPR}.
\bibitem[{Zhang et~al.(2021b)Zhang, Liu, Wang, Li and Hu}]{zhang2021learn}
\bibinfo{author}{Zhang, Z.}, \bibinfo{author}{Liu, Y.}, \bibinfo{author}{Wang,
  X.}, \bibinfo{author}{Li, B.}, \bibinfo{author}{Hu, W.},
  \bibinfo{year}{2021}b.
\newblock \bibinfo{title}{Learn to match: Automatic matching network design for
  visual tracking}, in: \bibinfo{booktitle}{ICCV}.
\bibitem[{Zhang et~al.(2020)Zhang, Peng, Fu, Li and Hu}]{Ocean_2020_ECCV}
\bibinfo{author}{Zhang, Z.}, \bibinfo{author}{Peng, H.}, \bibinfo{author}{Fu,
  J.}, \bibinfo{author}{Li, B.}, \bibinfo{author}{Hu, W.},
  \bibinfo{year}{2020}.
\newblock \bibinfo{title}{Ocean: Object-aware anchor-free tracking}, in:
  \bibinfo{booktitle}{ECCV}.

\end{thebibliography}
}

\newpage

\appendix

\section*{Appendix}

\section{Positional Encoding}
Transformer requires a positional encoding to identify the position of the current processing token~\cite{vaswani2017attention}. Through a series of comparison experiments, we choose \emph{untied positional encoding}, which is proposed in TUPE~\cite{ke2021rethinking}, as the positional encoding solution of our tracker. In addition, we generalize the \emph{untied positional encoding} to arbitrary dimensions to fit with other components in our tracker.


The original transformer~\cite{vaswani2017attention} proposes a absolute positional encoding method to represent the position: a fixed or learnable vector $p_i$ is assigned to each position $i$. Starting from the basic attention module, we have:
\begin{equation}
{\rm Atten}(Q,K,V)={\rm softmax} \Big(\frac{QK^T}{\sqrt{d_k}}V \Big),
\label{attn}
\end{equation}
where $Q$,$K$,$V$ are the $query$ vector, $key$ vector and $value$ vector, which are the parameters of the attention function, $d_k$ is the dimension of $key$. Introducing the linear projection matrix and multi-head attention to the attention module (\ref{attn}), we get the multi-head variant defined in \cite{vaswani2017attention}:
\begin{equation}
{\rm MultiHead}(Q,K,V) ={\rm Concat}({\rm head_1},...,{\rm head_h})W_O, 
\end{equation}
where ${\rm head_i} = {\rm Atten}(QW^Q_i,KW^K_i,VW^V_i)$, $W^Q_i\in\mathbb{R}^{d_{{\rm model}}\times d_k}$, $W^K_i\in\mathbb{R}^{d_{{\rm model}}\times d_k}$, $W^V_i\in\mathbb{R}^{d_{{\rm model}}\times d_v}$, $W^O_i\in\mathbb{R}^{hd_v\times d_{\rm model}}$ and $h$ is the number of heads.
For simplicity, as in \cite{ke2021rethinking}, we assume that $d_k=d_v=d_{\rm model}$, and use the single-head version of self-attention module. Denoting the input sequence as $x={x_1, x_2, \dots, x_n}$, where $n$ is the length of sequence, $x_i$ is the $i$-th token in the input data. Denoting the output sequence as $z=(z_1, z_2, \dots, z_n)$. Self-attention module can be rewritten as
\begin{align}
z_i&=\sum_{j=1}^n\frac{{\rm exp}(\alpha_{ij})}{\sum_{j'=1}^n {\rm exp}(\alpha_{ij'})}(x_jW^V), \\
{\rm where}\ \alpha_{ij}&=\frac{1}{\sqrt{d}}(x_iW^Q)(x_jW^K)^T.\label{attn_mat_elem}
\end{align}

Obviously, the self-attention module is permutation-invariance. Thus it can not "understand" the order of input tokens.

{\bf Untied absolute positional encoding.} By adding a learnable positional encoding~\cite{vaswani2017attention} to the single-head self-attention module, we can obtain the following equation:
\begin{equation}
\begin{split}
    \alpha_{ij}^{Abs}&=\frac{((w_i+p_i)W^Q)((w_j+p_j)W^K)^T}{\sqrt{d}} \\
                     &=\frac{(w_iW^Q)(w_jW^K)^T}{\sqrt{d}} + \frac{(w_iW^Q)(p_jW^K)^T}{\sqrt{d}} \\
                     &+\frac{(p_iW^Q)(w_jW^K)^T}{\sqrt{d}} + \frac{(p_iW^Q)(p_jW^K)^T}{\sqrt{d}}.\label{attn_mat_elem_with_pos}
\end{split}
\end{equation}

The equation (\ref{attn_mat_elem_with_pos}) is expanded into four terms: token-to-token, token-to-position, position-to-token, position-to-position. \cite{ke2021rethinking} discuss the problems that exist in the equation and proposes the \emph{untied absolute positional encoding}, which unties the correlation between tokens and positions by removing the token-position correlation terms in equation (\ref{attn_mat_elem_with_pos}), and using an isolated pair of projection matrices $U^Q$ and $U^K$ to perform linear transformation upon positional embedding vector. The following is the new formula for obtaining $\alpha_{ij}$ using the \emph{untied absolute positional encoding} in the $l$-th layer:
\begin{equation}
\begin{split}
    \alpha_{ij}&=\frac{1}{\sqrt{2d}}(x_i^lW^{Q,l})(x_j^lW^{K,l})^T \\
               &+\frac{1}{\sqrt{2d}}(p_iU^Q)(p_jU^K)^T.\label{untied_abs_pos_enc}
\end{split}
\end{equation}
where $p_i$ and $p_j$ is the positional embedding at position $i$ and $j$ respectively, $U^Q\in \mathbb{R}^{d\times d}$ and $U^K\in \mathbb{R}^{d\times d}$ are learnable projection matrices for the positional embedding vector. When extending to the multi-head version, the positional embedding $p_i$ is shared across different heads, while $U^Q$ and $U^K$ are different for each head. 

{\bf Relative positional bias.} According to \cite{shaw2018relpos}, relative positional encoding is a necessary supplement to absolute positional encoding. In~\cite{ke2021rethinking}, a relative positional encoding is applied by adding a relative positional bias to equation (\ref{untied_abs_pos_enc}):
\begin{equation}
\begin{split}
    \alpha_{ij}&=\frac{1}{\sqrt{2d}}(x_i^lW^{Q,l})(x_j^lW^{K,l})^T \\
              &+\frac{1}{\sqrt{2d}}(p_iU^Q)(p_jU^K)^T+b_{j-i},\label{untied_abs_pos_enc_with_rel}
\end{split}
\end{equation}
where for each $j-i$, $b_{j-i}$ is a learnable scalar. The \emph{relative positional bias} is also shared across layers. When extending to the multi-head version, $b_{j-i}$ is different for each head.

{\bf Generalize to multiple dimensions.} Before working with our tracker's encoder and decoder network, we need to extend the \emph{untied positional encoding} to a multi-dimensional version. One straightforward method is allocating a positional embedding matrix for every dimension and summing up all embedding vectors from different dimensions at the corresponding index to represent the final embedding vector. Together with \emph{relative positional bias}, for an $\mathtt{n}$-dimensional case, we have:
\begin{equation}
\begin{split}
    \alpha_{\underbrace{ij\dots}_\mathtt{n}, \underbrace{\vphantom{j}mn\dots}_\mathtt{n}}&=\frac{1}{\sqrt{2d}}(x_{\underbrace{ij\dots}_\mathtt{n}}W^Q)(x_{\underbrace{\vphantom{j}mn\dots}_\mathtt{n}}W^K)^T \\
               & +\frac{1}{\sqrt{2d}}[\underbrace{(p_i^1+p_j^2+\dots)}_\mathtt{n}U^Q] 
                [\underbrace{(p_m^1+p_n^2+\dots)}_\mathtt{n}U^K]^T \\
               & +b_{\underbrace{m-i,n-j, \dots}_\mathtt{n}} ~.
\end{split} 
\end{equation}

{\bf Generalize to concatenation-based fusion.} In order to work with \emph{concatenation-based fusion}, the \emph{untied absolute positional encoding} is also concatenated to match the real position, the indexing tuple of \emph{relative positional bias} now appends with a pair of indices to reflect the origination of $query$ and $key$ involved currently.

Take $l$-th layer in the encoder as the example:
\begin{equation}
\begin{split}
    \alpha_{ij,mn,g,h} &= \frac{1}{\sqrt{2d}}(x_{ij,g}^lW^{Q,l})(x_{mn,h}^lW^{K,l})^T \\
                      & +\frac{1}{\sqrt{2d}}[(p_{i,g}^1+p_{j,g}^2)U^Q_g] 
                       [(p_{m,h}^1+p_{n,h}^2)U^K_h]^T \\
                      & +b_{m-i,n-j,g,h}~,
\end{split}
\end{equation}
where $g$ and $h$ are the index of the origination of $query$ and $key$ respectively, for instance, $1$ for the tokens from the template image, $2$ for the tokens from the search image. The form in the decoder is similar, except that $g$ is fixed. In our implementation, the parameters of \emph{untied positional encoding} are shared inside the encoder and the decoder, respectively.

\section{The Effect of Pre-training Datasets}
The two variants of our tracker, SwinTrack-T-224 and SwinTrack-B-384 are using different pre-training datasets, which are derived from the settings from Swin Transformer~\cite{liu2021Swin}. Specifically, SwinTrack-T-224 adopts ImageNet-1k and SwinTrack-B-384 adopts ImageNet-22k. 

To analyze the effect of different pre-training datasets, we conduct an experiment on the performance of our tracker with different pre-training datasets. Other than the pre-training datasets, The experiment follows the same settings in the ablation study in the paper, the motion token is not used and the results on GOT-10k are trained on the full datasets as described in the paper. From Tab.~\ref{pretrain_compare}, we can observe that, for smaller model SwinTrack-T-224 (23M \# parameters), pre-training on ImageNet-22k brings small improvements on LaSOT (+0.6\%) and TrackingNet (+0.4\%) but degrades the performance on GOT-10k (-1.4\%). For larger model SwinTrack-B-384 (91M \# parameters), pre-training on ImageNet-22k shows significant performance gains on LaSOT (+2.2\%) and GOT-10k (+3.0\%) but slightly degrades the result on TrackingNet (-0.6\%). On LaSOT$_{\text{ext}}$, ImageNet-22k shows a performance degradation on smaller model SwinTrack-T-224 (-0.9\%) and brings small improvements on larger model SwinTrack-B-384 (+0.2\%).

\begin{table}[t]
\caption{The effect of Imagenet-22k pre-training. The results are following the settings in the ablation study in the paper (motion token is not used and the result on GOT-10k is trained on the full dataset).}
\label{pretrain_compare}
\begin{center}
\resizebox{\linewidth}{!}{
\begin{tabular}{r c cc cc cc ccc}
\hline
\multirow{2}{*}{Trackers} &
\multirow{2}{*}{Pre-training} &
\multicolumn{2}{c}{LaSOT~\cite{fan2019lasot}} & \multicolumn{2}{c}{LaSOT$_{\text{ext}}$~\cite{fan2021lasot}} & \multicolumn{2}{c}{TrackingNet~\cite{muller2018trackingnet}} & \multicolumn{3}{c}{GOT-10k~\cite{Huang2021got10k}} \\
&   & SUC & P    & SUC  & P   & SUC  & P  & AO    & SR$_{0.5}$ & SR$_{0.75}$ \\ \hline  \hline
SwinTrack-T-224 & ImageNet-1k  & 66.7    & 70.6   & 46.9 & 52.9 & 86.7   & 80.1 & 69.7 & 79.0   & 65.6   \\
SwinTrack-T-224 & ImageNet-22k & 67.3    & 71.7   & 46.0 & 51.7    & 81.2   & 78.9 & 69.5 & 78.9   & 65.5   \\
SwinTrack-B-384 & ImageNet-1k  & 68.0    & 72.5   & 47.3 & 53.2    & 83.8   & 82.9 & 71.8 & 80.2   & 67.1   \\
SwinTrack-B-384 & ImageNet-22k & 70.2    & 75.3   & 47.5 & 53.3    & 86.9   & 80.1 & 70.2 & 80.7   & 65.4   \\
\hline
\end{tabular}}
\end{center}
\end{table}

\begin{table}[t]
\caption{Performance comparisons with newly released Transformer-based Trackers on four benchmarks: LaSOT, LaSOT$_{\text{ext}}$, TrackingNet and GOT-10k.}
\label{sota_compare_extra}
\begin{center}
\resizebox{\linewidth}{!}{
\begin{tabular}{r l cc cc cc ccc cc}
\hline
\multirow{2}{*}{Tracker} &
\multirow{2}{*}{Pre-training} &
  \multicolumn{2}{c}{LaSOT~\cite{fan2019lasot}} &
  \multicolumn{2}{c}{LaSOT$_{\text{ext}}$~\cite{fan2021lasot}} &
  \multicolumn{2}{c}{TrackingNet~\cite{muller2018trackingnet}} &
  \multicolumn{3}{c}{GOT-10k~\cite{Huang2021got10k}} \\
                                 &       & SUC  & P    & SUC  & P    & SUC  & P    & AO   & SR$_{0.5}$ & SR$_{0.75}$ \\ \hline \hline
STARK~\cite{stark}                & ImageNet-1k      & 67.1 & -    & -    & -    & 82.0 & -    & 68.8 & 78.1       & 64.1         \\ 
SBT~\cite{xie2022correlation}     & ImageNet-1k      & 66.7 & 71.1 & -    & -    & -    & -    & 70.4 & 80.8       & 64.7          \\ 
ToMP~\cite{mayer2022transforming} & ImageNet-1k      & 68.5 & 73.5 & 45.9 & -    & 81.5 & 78.9 & -    & -          & -            \\ 
MixFormer~\cite{cui2022mixformer} & ImageNet-22k      & 70.1 & 76.3 & -    & -    & 83.9 & 83.1 & -    & -          & -           \\ 
AiATrack~\cite{gao2022aiatrack}   & ImageNet-1k      & 69.0 & 73.8 & 47.7 & 55.4 & 82.7 & 80.4 & 69.6 & 80.0       & 63.2        \\ 
Unicorn~\cite{unicorn}            & ImageNet-1k      & 68.5 & 74.1 & -    & -    & 83.0 & 82.2 & -    & -          & -            \\ 
OSTrack~\cite{ye2022ostrack}      & MAE~\cite{he2022masked} & 71.1 & 77.6 & 50.5    & 57.6 & 83.9 & 83.2  & 73.7 & 83.2 & 70.8     \\ 
\hline

SwinTrack-T-224                   & ImageNet-1k      & 67.2 & 70.8 & 47.6 & 53.9 & 81.1 & 78.4 & 71.3 & 81.9       & 64.5       \\
SwinTrack-B-384                   & ImageNet-22k      & 71.3 & 76.5 & 49.1 & 55.6 & 84.0 & 82.8 & 72.4 & 80.5       & 67.8       \\
\hline
 \end{tabular}
}
\end{center}
\end{table}

\section{Comparison with Newly Released Transformer-based Trackers}
We compare our tracker with some newly released Transformer-based trackers, including STARK~\cite{stark}, SBT~\cite{xie2022correlation}, ToMP~\cite{mayer2022transforming}, MixFormer~\cite{cui2022mixformer}, AiATrack~\cite{gao2022aiatrack}, Unicorn~\cite{unicorn}, OSTrack~\cite{ye2022ostrack} in Tab.~\ref{sota_compare_extra} in four challenging benchmarks. The result shows our tracker is still competitive.

Fig.~\ref{success_fig} and Fig.~\ref{precision_fig} show the success plot and the precision plot respectively. The comparison includes our SwinTrack-T-224, our SwinTrack-B-384, TransT\cite{TransT}, STARK\cite{stark}, MixFormer\cite{cui2022mixformer}, AiATrack\cite{gao2022aiatrack} and ToMP\cite{mayer2022transforming}. Our tracker obtained the best performance on this benchmark. By looking into the curves of the figures, there is a significant advantage in the bounding box accuracy compared with other trackers due to our fully attentional architecture.

\begin{figure}
     \centering
     \caption{Comparison to the state-of-the-art trackers using success (SUC) AUC score under different attributes of LaSOT~\cite{fan2019lasot} Test set.}
    \label{main_attr_suc_fig}
     \centering
    \includegraphics[width=\linewidth]{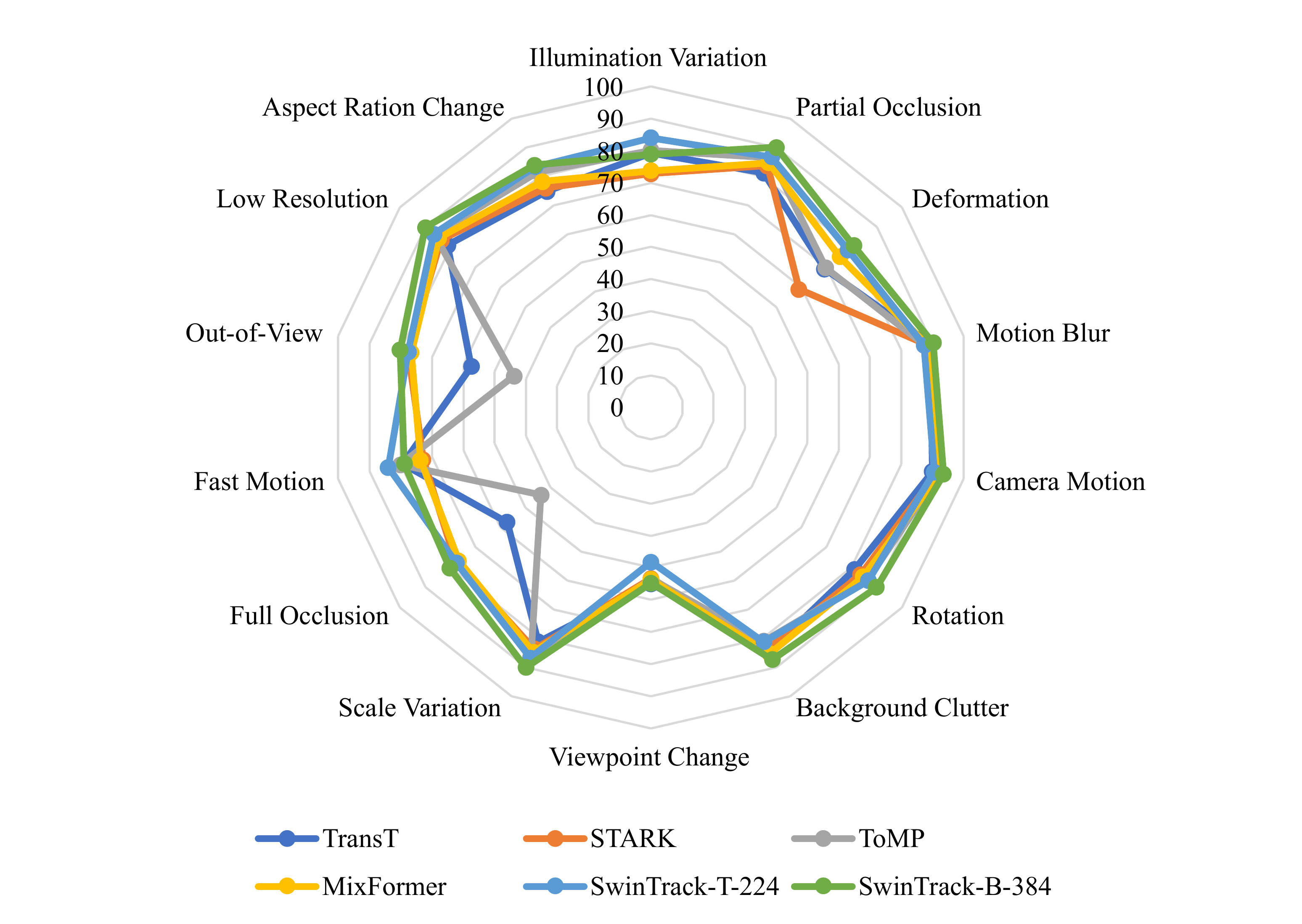}
\end{figure}

The success AUC score under different attributes of LaSOT~\cite{fan2019lasot} Test set in shown in Fig.~\ref{main_attr_suc_fig}. Fig.~\ref{main_attr_suc_fig} indicates that our tracker has no obvious shortcomings except the viewpoint change.

\section{Results on UAV123 and VOT Benchmark}

In this section, we report the performance of the tracker on three additional benchmarks, including UAV123~\cite{mueller2016benchmark}, VOT2020 and VOT-STB2022~\cite{VOT_TPAMI}.

UAV123\cite{mueller2016benchmark} is an aerial video dataset and benchmark for low-altitude UAV target tracking, containing 123 video sequences. Our tracker is on par with the state-of-the-art, AiATrack~\cite{gao2022aiatrack}, on this benchmark. The results are shown in Tab.~\ref{uav_compare}. 

Finally, we evaluate our tracker on the two versions of the VOT Challenge: VOT2020 and VOT-STB2022. The VOT2020 dataset contains 60 videos with segmentation masks annotated. Since our tracker is a bounding box only method, we compare the results with the trackers that produce the bounding boxes as well. The result in Tab.\ref{vot2020compare} shows that SwinTrack-T-224 has a better performance than the larger SwinTrack-B-384 on this benchmark.

In addition, We report the results on VOT-STB2022 in Tab.\ref{vot2022compare}. SwinTrack-T-224 has a better performance on VOT-STB2022 as well. No comparison is made since VOT-STB2022 is a newly released benchmark.

\begin{table}[t]
\caption{Comparison to the state-of-the-arts on UAV123~\cite{mueller2016benchmark} benchmark.}
\label{uav_compare}
\begin{center}
\resizebox{\linewidth}{!}{
\begin{tabular}{c cccccccccc cc}
\hline
 & \begin{tabular}[c]{@{}c@{}}Ocean\\ \cite{Ocean_2020_ECCV}\end{tabular}
 & \begin{tabular}[c]{@{}c@{}}DiMP\\ \cite{bhat2019learning}\end{tabular}
 & \begin{tabular}[c]{@{}c@{}}TransT\\ \cite{TransT}\end{tabular}
 & \begin{tabular}[c]{@{}c@{}}ToMP\\ 50\cite{mayer2022transforming}\end{tabular}
 & \begin{tabular}[c]{@{}c@{}}MixFormer\\ 22k\cite{cui2022mixformer}\end{tabular}
  & \begin{tabular}[c]{@{}c@{}}AiATrack\\ \cite{gao2022aiatrack}\end{tabular}
 
 & \begin{tabular}[c]{@{}c@{}}SwinTrack\\T-224\end{tabular}
 & \begin{tabular}[c]{@{}c@{}}SwinTrack\\B-384\end{tabular}
 \\ \hline
AUC (\%) & 62.1 & 65.3 & 69.1 & 69.0 & 70.4 & 70.6
& 68.8 & 70.5 \\ \hline

\end{tabular}}
\end{center}
\end{table}

\begin{table}[t]
\caption{Comparison to the state-of-the-art bounding box only methods on VOT2020ST~\cite{VOT_TPAMI}.}
\label{vot2020compare}
\begin{center}
\resizebox{\linewidth}{!}{
\begin{tabular}{l cccccccccc cc}
\hline
 & \begin{tabular}[c]{@{}c@{}}ATOM\\ \cite{danelljan2019atom}\end{tabular}
 & \begin{tabular}[c]{@{}c@{}}DiMP\\ \cite{bhat2019learning}\end{tabular}
 & \begin{tabular}[c]{@{}c@{}}STARK\\ 50\cite{stark}\end{tabular}
 & \begin{tabular}[c]{@{}c@{}}STARK\\ 101\cite{stark}\end{tabular}
 & \begin{tabular}[c]{@{}c@{}}ToMP\\ 50\cite{mayer2022transforming}\end{tabular}
  & \begin{tabular}[c]{@{}c@{}}ToMP\\ 101\cite{mayer2022transforming}\end{tabular}
 
 & \begin{tabular}[c]{@{}c@{}}SwinTrack\\T-224\end{tabular}
 & \begin{tabular}[c]{@{}c@{}}SwinTrack\\B-384\end{tabular}
 \\ \hline
EAO           & 0.271 & 0.274 & 0.308 & 0.303 & 0.297 & 0.309
& 0.302 & 0.283 \\ 
Accuracy & 0.462 & 0.457 & 0.478 & 0.481 & 0.453 & 0.453
& 0.471 & 0.472 \\ 
Robustness & 0.734 & 0.734 & 0.799 & 0.775 & 0.789 & 0.814
& 0.775 & 0.741 \\ \hline

\end{tabular}}
\end{center}
\end{table}

\begin{table}[t]
\caption{Results on VOT-STB2022~\cite{VOT_TPAMI}.}
\label{vot2022compare}
\begin{center}
\begin{tabular}{l cccccccccc cc}
\hline
 & \begin{tabular}[c]{@{}c@{}}SwinTrack\\T-224\end{tabular}
 & \begin{tabular}[c]{@{}c@{}}SwinTrack\\B-384\end{tabular}
 \\ \hline
EAO        & 0.505 & 0.477 \\ 
Accuracy   & 0.777 & 0.790 \\ 
Robustness & 0.790 & 0.759 \\
\hline
\end{tabular}
\end{center}
\end{table}

\section{Quantitative Analysis of the Effectiveness of Motion Token}

\begin{figure}
    \caption{Success (SUC) AUC score on LaSOT~\cite{fan2019lasot} Test set assessing the effectiveness of the motion token.}
    \label{motion_token_suc_fig}
    \centering
\begin{tikzpicture}

\definecolor{darkgray176}{RGB}{176,176,176}

\begin{axis}[
legend cell align={left},
legend style={fill opacity=1.0, draw opacity=1, text opacity=1, at={(0.03,0.03)}, anchor=south west},
tick align=outside,
tick pos=left,
title={Success plot},
x grid style={darkgray176},
xlabel={Overlap threshold},
xmajorgrids,
xmin=0, xmax=1,
xtick style={color=black},
y grid style={darkgray176},
ylabel={Overlap Precision [\%]},
ymajorgrids,
ymin=0, ymax=100,
ytick style={color=black}
]
\addplot [thick, black]
table {%
0 85.3124008178711
0.0500000007450581 83.784065246582
0.100000001490116 82.8986511230469
0.150000005960464 82.199348449707
0.200000002980232 81.6311187744141
0.25 81.1087036132812
0.300000011920929 80.5525436401367
0.349999994039536 79.9055252075195
0.400000005960464 79.166862487793
0.449999988079071 78.3173751831055
0.5 77.1825637817383
0.550000011920929 75.8297119140625
0.600000023841858 74.2671813964844
0.649999976158142 72.1601104736328
0.699999988079071 69.3161926269531
0.75 65.2747192382812
0.800000011920929 59.0510787963867
0.850000023841858 49.2461814880371
0.899999976158142 33.1640968322754
0.949999988079071 10.2892179489136
1 0
};
\addlegendentry{SwinTrack-T-224-NoMToken [66.7]}
\addplot [thick, blue]
table {%
0 85.5778732299805
0.0500000007450581 84.1035537719727
0.100000001490116 83.3299713134766
0.150000005960464 82.7461700439453
0.200000002980232 82.2377166748047
0.25 81.672737121582
0.300000011920929 81.1188430786133
0.349999994039536 80.4158172607422
0.400000005960464 79.6402816772461
0.449999988079071 78.8200378417969
0.5 77.7748489379883
0.550000011920929 76.5463180541992
0.600000023841858 75.0328216552734
0.649999976158142 72.9672393798828
0.699999988079071 70.0464401245117
0.75 65.843864440918
0.800000011920929 59.4688110351562
0.850000023841858 49.6399803161621
0.899999976158142 33.6708526611328
0.949999988079071 10.877124786377
1 0
};
\addlegendentry{SwinTrack-T-224 [67.2]}
\addplot [thick, green]
table {%
0 85.850227355957
0.0500000007450581 84.7379379272461
0.100000001490116 84.1397018432617
0.150000005960464 83.7286987304688
0.200000002980232 83.3811416625977
0.25 82.9725036621094
0.300000011920929 82.5358123779297
0.349999994039536 81.955322265625
0.400000005960464 81.282958984375
0.449999988079071 80.5210647583008
0.5 79.7132339477539
0.550000011920929 78.7315979003906
0.600000023841858 77.4594192504883
0.649999976158142 75.8076171875
0.699999988079071 73.6182556152344
0.75 70.5351638793945
0.800000011920929 65.7851028442383
0.850000023841858 57.7718811035156
0.899999976158142 43.9315452575684
0.949999988079071 19.8706531524658
1 0
};
\addlegendentry{SwinTrack-B-384-NoMToken [70.2]}
\addplot [thick, red]
table {%
0 87.1718978881836
0.0500000007450581 86.1381683349609
0.100000001490116 85.4462280273438
0.150000005960464 84.9568328857422
0.200000002980232 84.5217132568359
0.25 84.090202331543
0.300000011920929 83.6422119140625
0.349999994039536 83.1625366210938
0.400000005960464 82.4673690795898
0.449999988079071 81.7340850830078
0.5 80.963996887207
0.550000011920929 79.9329605102539
0.600000023841858 78.6637268066406
0.649999976158142 77.0235595703125
0.699999988079071 74.7673721313477
0.75 71.6431579589844
0.800000011920929 66.8184432983398
0.850000023841858 58.8123397827148
0.899999976158142 44.7114028930664
0.949999988079071 20.9500827789307
1 0
};
\addlegendentry{SwinTrack-B-384 [71.3]}
\end{axis}

\end{tikzpicture}
\end{figure}
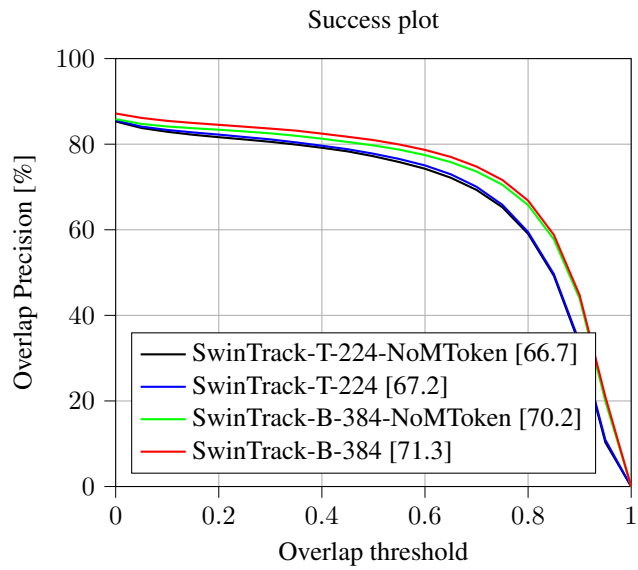

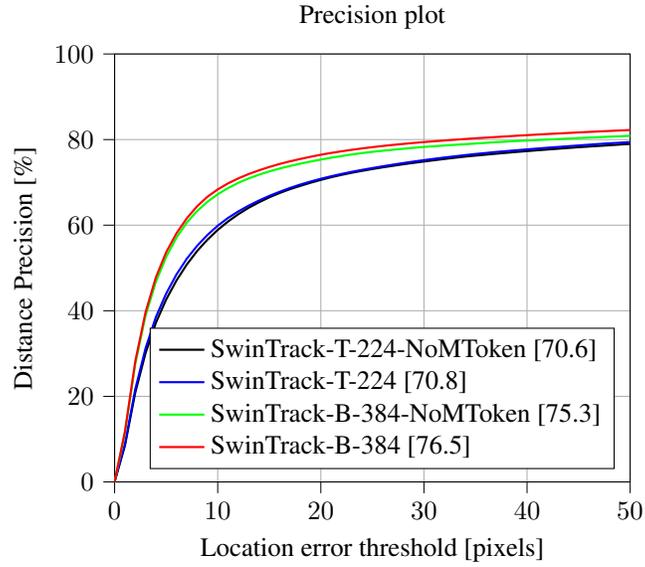
\begin{figure}
    \caption{Precision (PRE) AUC score on LaSOT~\cite{fan2019lasot} Test set assessing the effectiveness of the motion token.}
    \label{motion_token_pre_fig}
    \centering
\begin{tikzpicture}

\definecolor{darkgray176}{RGB}{176,176,176}

\begin{axis}[
legend cell align={left},
legend style={fill opacity=1.0, draw opacity=1, text opacity=1, at={(0.97,0.03)}, anchor=south east},
tick align=outside,
tick pos=left,
title={Precision plot},
x grid style={darkgray176},
xlabel={Location error threshold [pixels]},
xmajorgrids,
xmin=0, xmax=50,
xtick style={color=black},
y grid style={darkgray176},
ylabel={Distance Precision [\%]},
ymajorgrids,
ymin=0, ymax=100,
ytick style={color=black}
]
\addplot [thick, black]
table {%
0 0.0504663623869419
1 8.46163272857666
2 20.8333606719971
3 30.1140193939209
4 37.180606842041
5 42.6359939575195
6 47.0817680358887
7 50.8075790405273
8 53.9556846618652
9 56.5992660522461
10 58.8906402587891
11 60.8490180969238
12 62.5650177001953
13 64.0610122680664
14 65.3373031616211
15 66.500129699707
16 67.5050964355469
17 68.4029922485352
18 69.1713256835938
19 69.9066162109375
20 70.5755157470703
21 71.1993865966797
22 71.7446899414062
23 72.2436828613281
24 72.6890640258789
25 73.1140365600586
26 73.5050048828125
27 73.869010925293
28 74.235954284668
29 74.5569915771484
30 74.8698577880859
31 75.1739120483398
32 75.4675903320312
33 75.7525253295898
34 76.0183334350586
35 76.2531661987305
36 76.472282409668
37 76.6931762695312
38 76.9036407470703
39 77.1119537353516
40 77.3035202026367
41 77.4973220825195
42 77.6762542724609
43 77.8576583862305
44 78.0350570678711
45 78.2102203369141
46 78.366096496582
47 78.5194625854492
48 78.6722717285156
49 78.8311920166016
50 78.9822845458984
};
\addlegendentry{SwinTrack-T-224-NoMToken [70.6]}
\addplot [thick, blue]
table {%
0 0.0504663623869419
1 8.82082366943359
2 21.6383800506592
3 31.2220611572266
4 38.507209777832
5 44.0193634033203
6 48.4878005981445
7 52.1530570983887
8 55.1679916381836
9 57.7008743286133
10 59.8693313598633
11 61.6847267150879
12 63.2172889709473
13 64.5631484985352
14 65.7556457519531
15 66.8354949951172
16 67.7942886352539
17 68.6865234375
18 69.4827346801758
19 70.1895065307617
20 70.8372116088867
21 71.4245758056641
22 71.9685516357422
23 72.4641418457031
24 72.9245529174805
25 73.3448638916016
26 73.756721496582
27 74.1450805664062
28 74.5152206420898
29 74.8672561645508
30 75.1938400268555
31 75.5105285644531
32 75.8148727416992
33 76.0864944458008
34 76.3699798583984
35 76.6304931640625
36 76.8664627075195
37 77.1056671142578
38 77.3082580566406
39 77.505859375
40 77.7110290527344
41 77.9050674438477
42 78.0927734375
43 78.272087097168
44 78.4523544311523
45 78.6244125366211
46 78.8016357421875
47 78.9714889526367
48 79.127555847168
49 79.2715682983398
50 79.4226913452148
};
\addlegendentry{SwinTrack-T-224 [70.8]}
\addplot [thick, green]
table {%
0 0.0504663623869419
1 11.3507537841797
2 27.501615524292
3 38.8654556274414
4 46.7537078857422
5 52.5928611755371
6 57.0894165039062
7 60.5836219787598
8 63.3018035888672
9 65.4725646972656
10 67.2124557495117
11 68.6652221679688
12 69.8958129882812
13 70.9169082641602
14 71.788215637207
15 72.566650390625
16 73.2369613647461
17 73.8309860229492
18 74.3663482666016
19 74.8588943481445
20 75.326904296875
21 75.7556762695312
22 76.1479644775391
23 76.5143966674805
24 76.8414459228516
25 77.1409378051758
26 77.390983581543
27 77.6254043579102
28 77.8582305908203
29 78.0759735107422
30 78.2733535766602
31 78.4513931274414
32 78.6177368164062
33 78.7706298828125
34 78.9235687255859
35 79.0780944824219
36 79.2373123168945
37 79.3906860351562
38 79.5198516845703
39 79.6586990356445
40 79.7798461914062
41 79.890380859375
42 80.0062484741211
43 80.120002746582
44 80.234733581543
45 80.3514785766602
46 80.4606552124023
47 80.5631103515625
48 80.66796875
49 80.7707443237305
50 80.8708877563477
};
\addlegendentry{SwinTrack-B-384-NoMToken [75.3]}
\addplot [thick, red]
table {%
0 0.0504663623869419
1 11.8184518814087
2 28.1475982666016
3 39.6667289733887
4 47.7942314147949
5 53.6860885620117
6 58.1003837585449
7 61.6025276184082
8 64.4151992797852
9 66.6277313232422
10 68.3537902832031
11 69.7637023925781
12 70.930549621582
13 71.9451446533203
14 72.8109664916992
15 73.5948333740234
16 74.2950057983398
17 74.9093780517578
18 75.4595718383789
19 75.9770965576172
20 76.4579238891602
21 76.8763275146484
22 77.2583541870117
23 77.6085052490234
24 77.935188293457
25 78.2289581298828
26 78.5010604858398
27 78.7525863647461
28 78.9847564697266
29 79.2096710205078
30 79.4118957519531
31 79.6002960205078
32 79.7815246582031
33 79.9609222412109
34 80.1327056884766
35 80.3031997680664
36 80.45068359375
37 80.5997009277344
38 80.7397308349609
39 80.8952789306641
40 81.0318756103516
41 81.1686553955078
42 81.2992172241211
43 81.4214248657227
44 81.5451126098633
45 81.6689224243164
46 81.7865219116211
47 81.9021072387695
48 82.0111312866211
49 82.1211853027344
50 82.2361602783203
};
\addlegendentry{SwinTrack-B-384 [76.5]}
\end{axis}

\end{tikzpicture}
\end{figure}

\begin{figure}
    \caption{Success (SUC) AUC score under different attributes of LaSOT~\cite{fan2019lasot} Test set assessing the effectiveness of the motion token.}
    \label{motion_token_attribute_suc_fig}
    \centering
    \includegraphics[width=\linewidth]{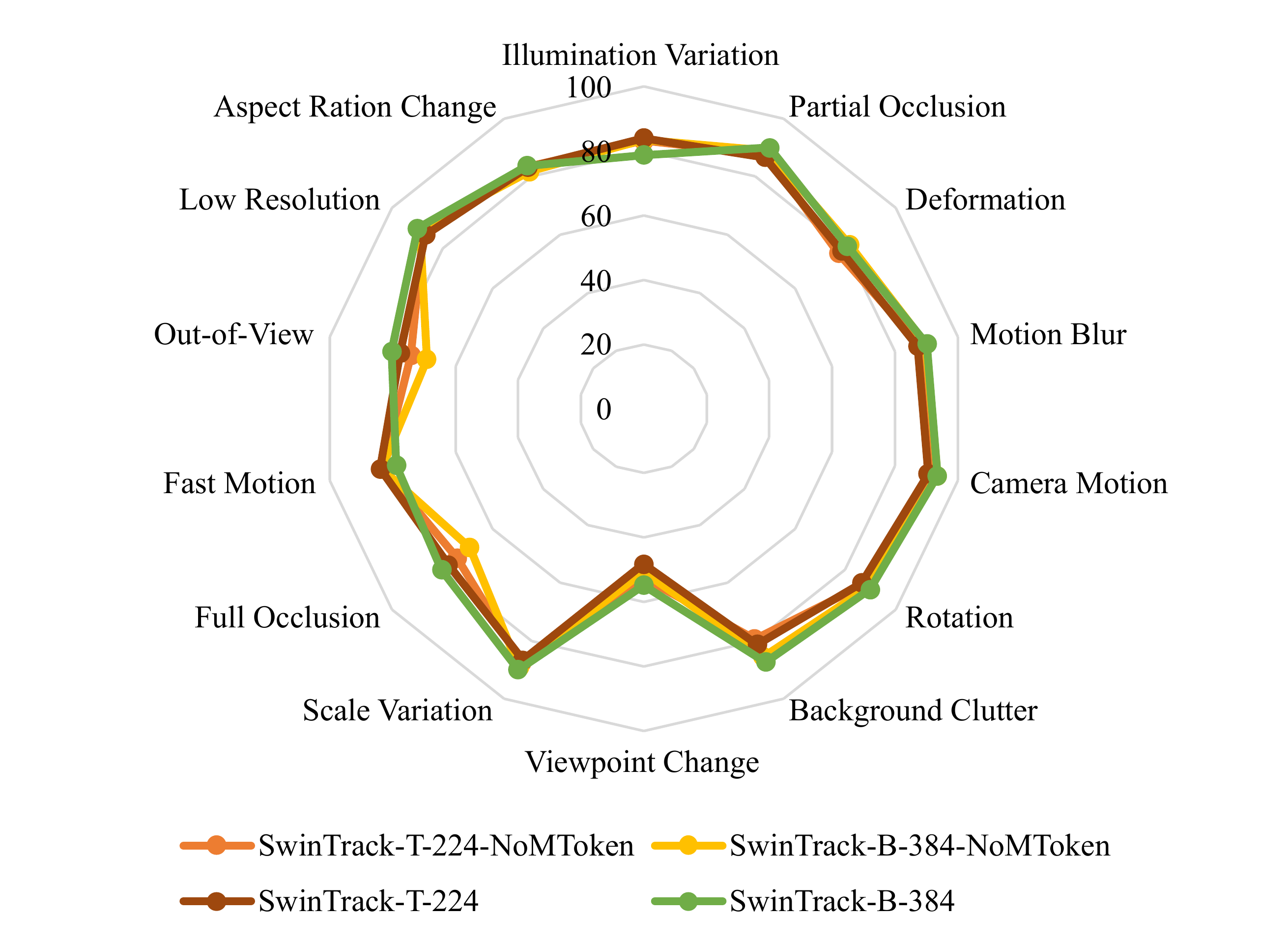}
\end{figure}

To give a further analysis of the effectiveness of motion token, we provide the success plot (Fig.~\ref{motion_token_suc_fig}) and precision plot (Fig.~\ref{motion_token_pre_fig}) on LaSOT test set, and the success AUC score under different attributes of LaSOT test set in Fig.~\ref{motion_token_attribute_suc_fig}. The success plot and the precision plot show that the motion token improves the performance of the trackers by boosting robustness. While the Fig.~\ref{motion_token_attribute_suc_fig} further points out that the motion token can assist the tracker to recover from a failure state when the vision features are not reliable like an object is getting out of view or fully occluded by other objects.

\section{Response Visualization for Qualitative Analysis}
\begin{figure}
\caption{Heatmap visualization of the tracking response map of our SwinTrack-B-384 on LaSOT$_{\mathrm{ext}}$~\cite{fan2021lasot}. The odd rows visualize the search region patches with ground-truth bounding box (in \HF{red} rectangles). The even rows visualize the search region patches blended with the heatmap visualization of the response map. The sequences and challenges involved: atv-10 (POC, ROT, VC, SV, LR, ARC), wingsuit-10 (CM, BC, VC, SV, FOC, LR, ARC), rhino-9 (DEF, SV, ARC) and misc-3 (POC, MB, ROT, BC, SV, FOC, FM, LR).}
\label{heatmap_viz}
\centering
\bgroup
\setlength{\tabcolsep}{1pt}
\begin{tabular}{ccccc}
\subfloat{\includegraphics[width = 0.15\textwidth]{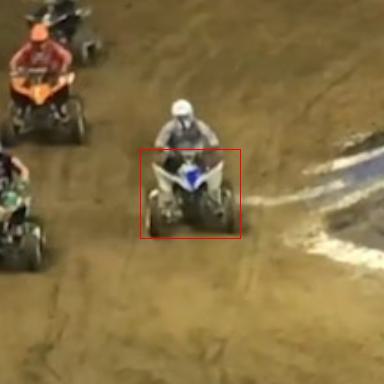}} &
\subfloat{\includegraphics[width = 0.15\textwidth]{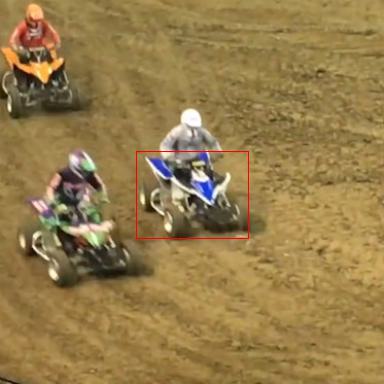}} &
\subfloat{\includegraphics[width = 0.15\textwidth]{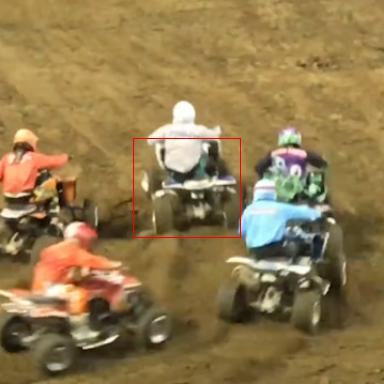}} &
\subfloat{\includegraphics[width = 0.15\textwidth]{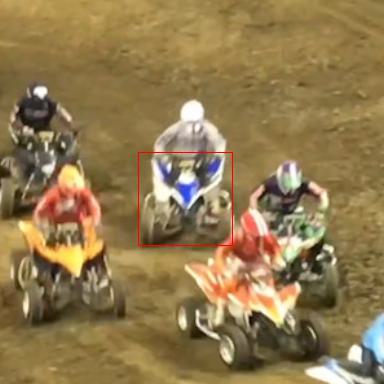}} &
\subfloat{\includegraphics[width = 0.15\textwidth]{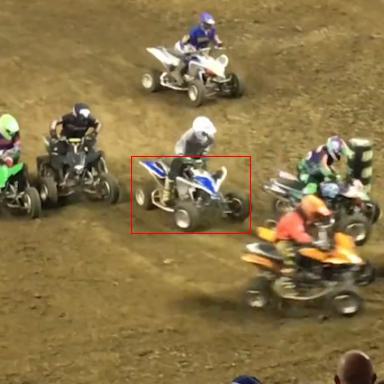}}\\
[-16pt]
\subfloat{\includegraphics[width = 0.15\textwidth]{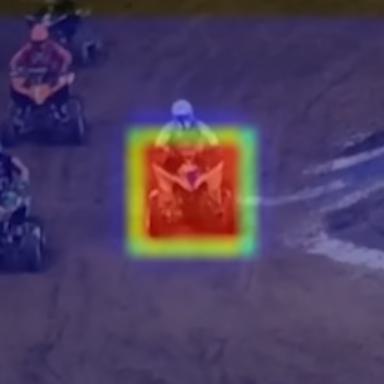}} &
\subfloat{\includegraphics[width = 0.15\textwidth]{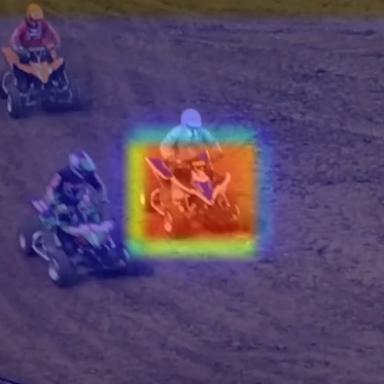}} &
\subfloat{\includegraphics[width = 0.15\textwidth]{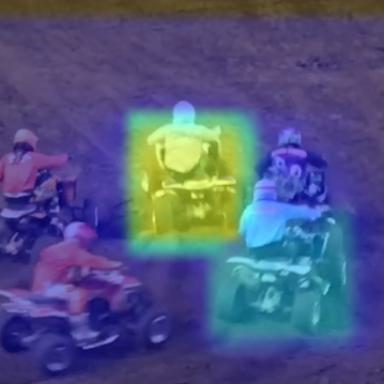}} &
\subfloat{\includegraphics[width = 0.15\textwidth]{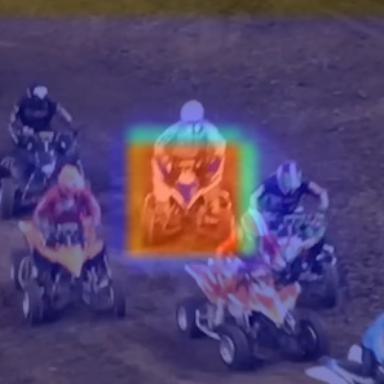}} &
\subfloat{\includegraphics[width = 0.15\textwidth]{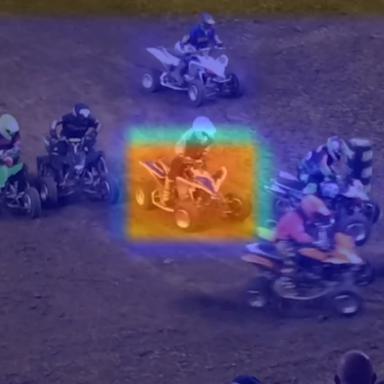}}\\
[-14pt]
\subfloat{\includegraphics[width = 0.15\textwidth]{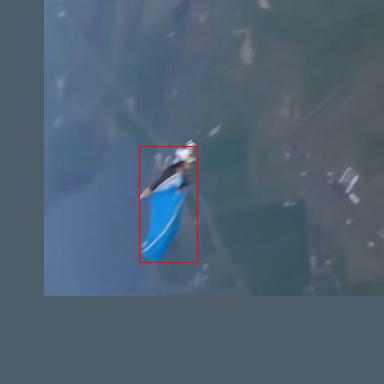}} &
\subfloat{\includegraphics[width = 0.15\textwidth]{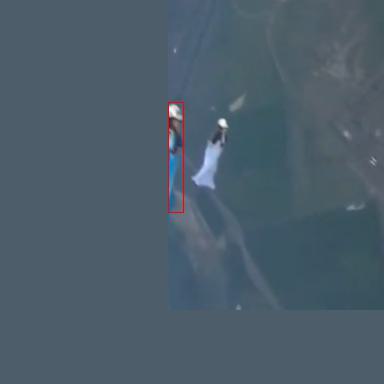}} &
\subfloat{\includegraphics[width = 0.15\textwidth]{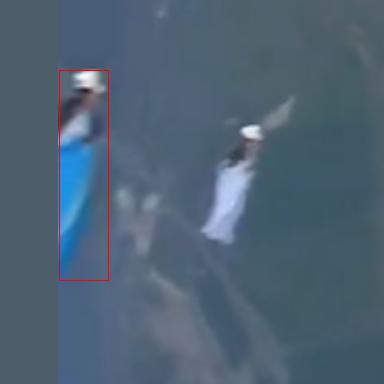}} &
\subfloat{\includegraphics[width = 0.15\textwidth]{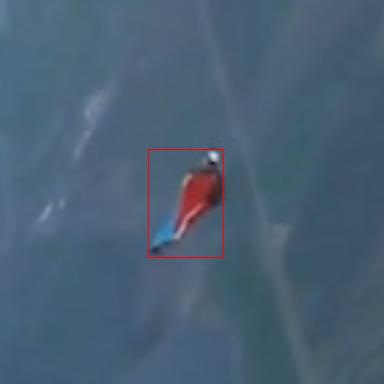}} &
\subfloat{\includegraphics[width = 0.15\textwidth]{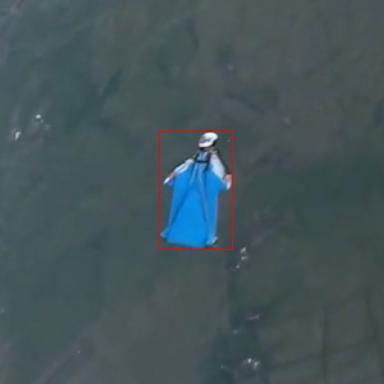}}\\
[-16pt]
\subfloat{\includegraphics[width = 0.15\textwidth]{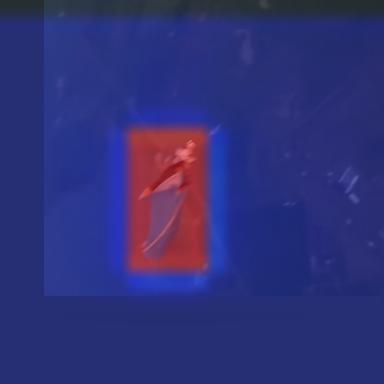}} &
\subfloat{\includegraphics[width = 0.15\textwidth]{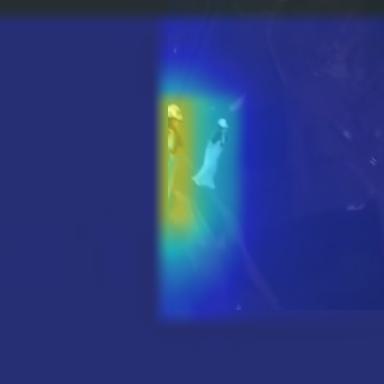}} &
\subfloat{\includegraphics[width = 0.15\textwidth]{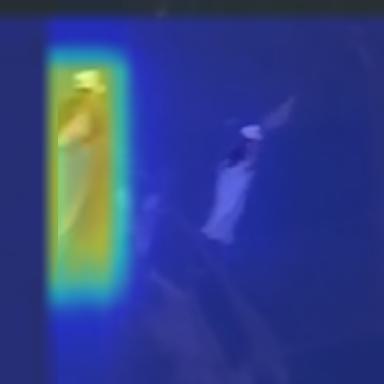}} &
\subfloat{\includegraphics[width = 0.15\textwidth]{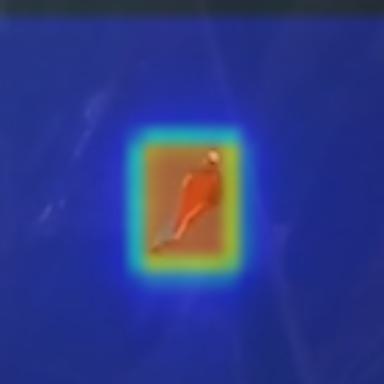}} &
\subfloat{\includegraphics[width = 0.15\textwidth]{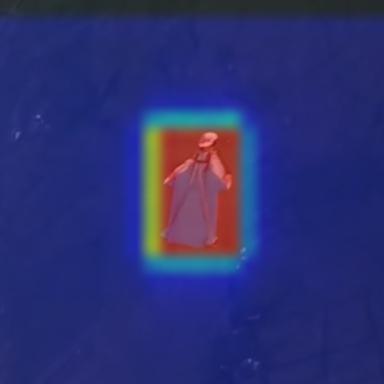}}\\
[-14pt]
\subfloat{\includegraphics[width = 0.15\textwidth]{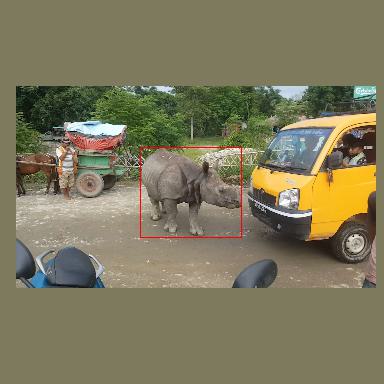}} &
\subfloat{\includegraphics[width = 0.15\textwidth]{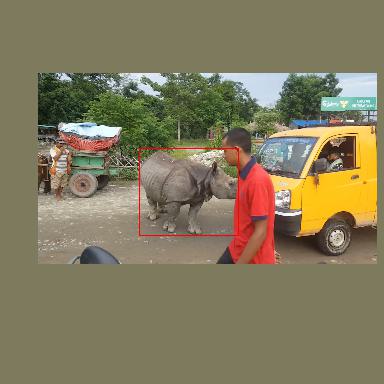}} &
\subfloat{\includegraphics[width = 0.15\textwidth]{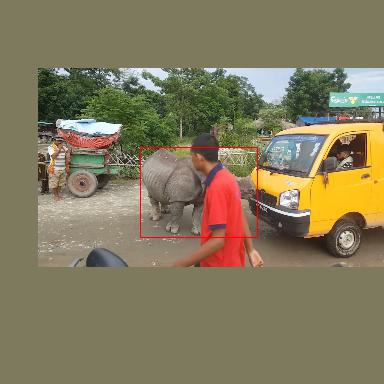}} &
\subfloat{\includegraphics[width = 0.15\textwidth]{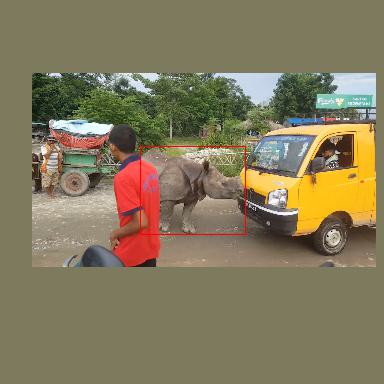}} &
\subfloat{\includegraphics[width = 0.15\textwidth]{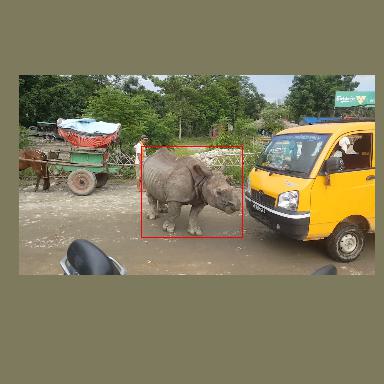}}\\
[-16pt]
\subfloat{\includegraphics[width = 0.15\textwidth]{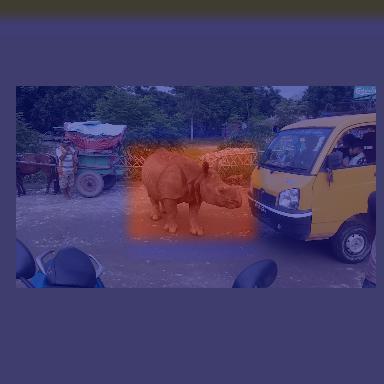}} &
\subfloat{\includegraphics[width = 0.15\textwidth]{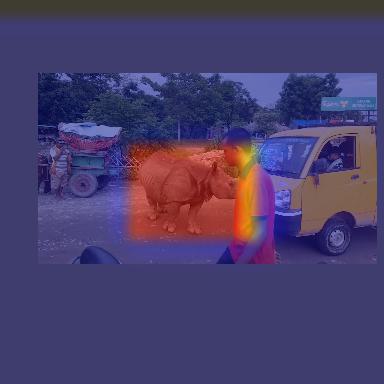}} &
\subfloat{\includegraphics[width = 0.15\textwidth]{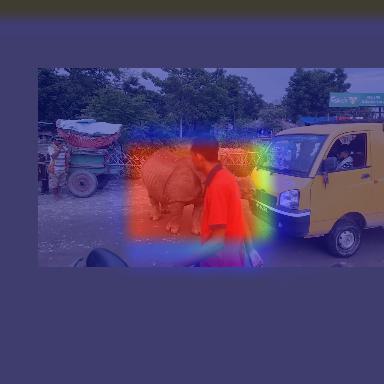}} &
\subfloat{\includegraphics[width = 0.15\textwidth]{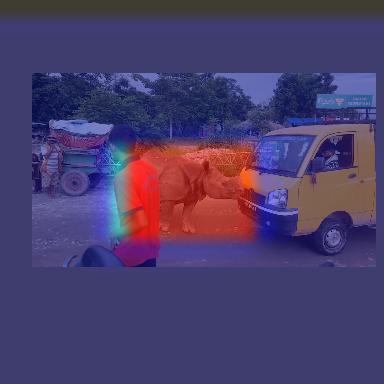}} &
\subfloat{\includegraphics[width = 0.15\textwidth]{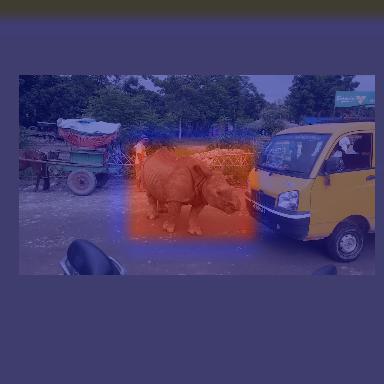}}\\
[-14pt]
\subfloat{\includegraphics[width = 0.15\textwidth]{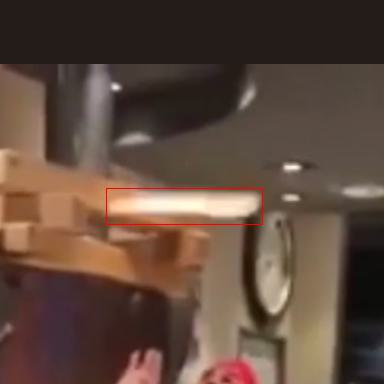}} &
\subfloat{\includegraphics[width = 0.15\textwidth]{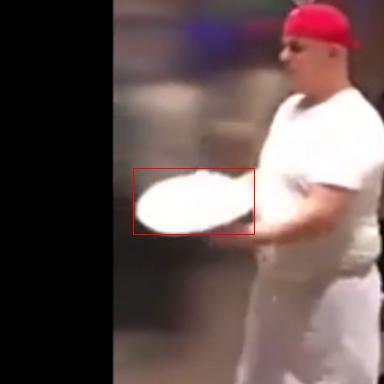}} &
\subfloat{\includegraphics[width = 0.15\textwidth]{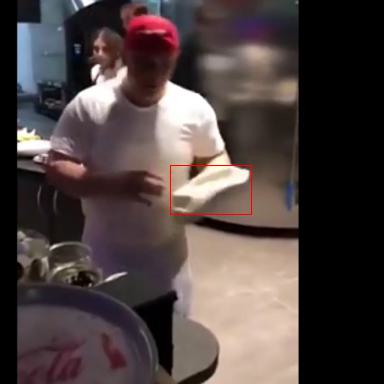}} &
\subfloat{\includegraphics[width = 0.15\textwidth]{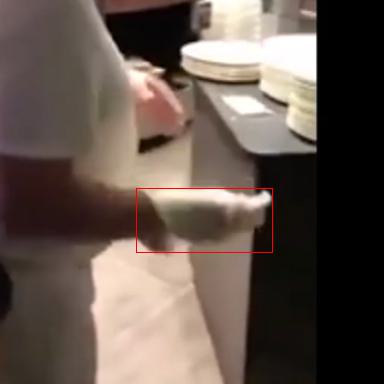}} &
\subfloat{\includegraphics[width = 0.15\textwidth]{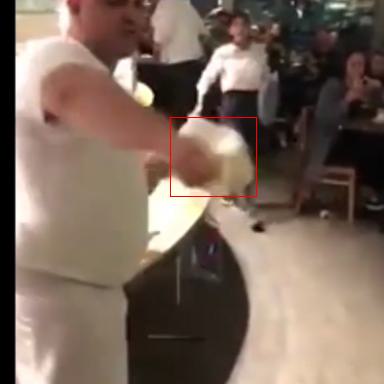}}\\
[-16pt]
\subfloat{\includegraphics[width = 0.15\textwidth]{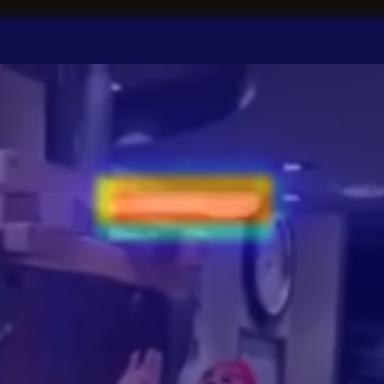}} &
\subfloat{\includegraphics[width = 0.15\textwidth]{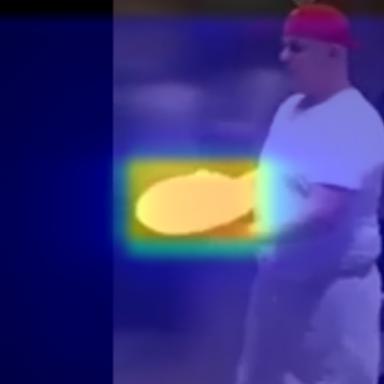}} &
\subfloat{\includegraphics[width = 0.15\textwidth]{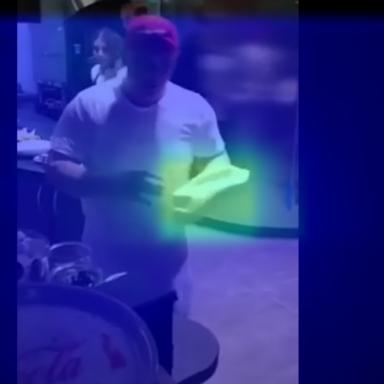}} &
\subfloat{\includegraphics[width = 0.15\textwidth]{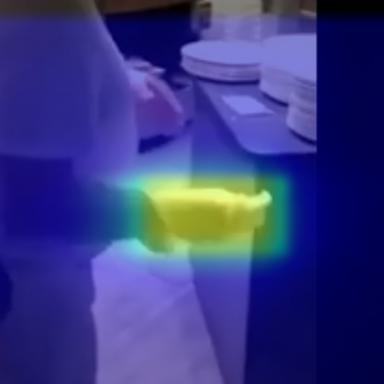}} &
\subfloat{\includegraphics[width = 0.15\textwidth]{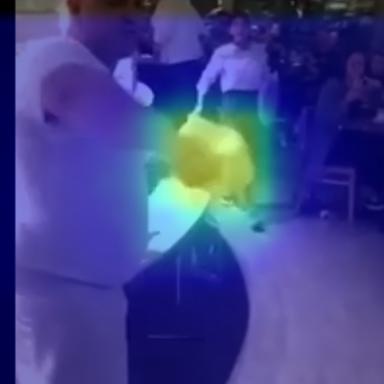}}
\end{tabular}
\egroup
\end{figure}

We provide the heatmap visualization of the response map generated by the IoU-aware classification branch of the head in our SwinTrack-B-384 in Fig.~\ref{heatmap_viz}. The visualized sequences are from LaSOT$_{\mathrm{ext}}$~\cite{fan2021lasot}, with challenges include fast motion, full occlusion, hard distractor, \etc. The results demonstrate the great discriminative power of our tracker. Many trackers will show a multi-peak on the response map when the target object is occluded or multiple similar objects exist. With the vision-motion integrated Transformer architecture, our tracker eases such phenomenon.

\section{Failure Case}
We show some typical failure cases of our tracker (SwinTrack-B-384 on LaSOT$_{\mathrm{ext}}$~\cite{fan2021lasot} and VOT-STB2022~\cite{VOT_TPAMI}) in Fig.~\ref{heatmap_fail_viz}. The first case suffers from a mixture of low resolution, fast motion, and background clutter. The second case suffers from a fast occlusion by a distractor. The third case suffers from the non-semantic target.

\newpage

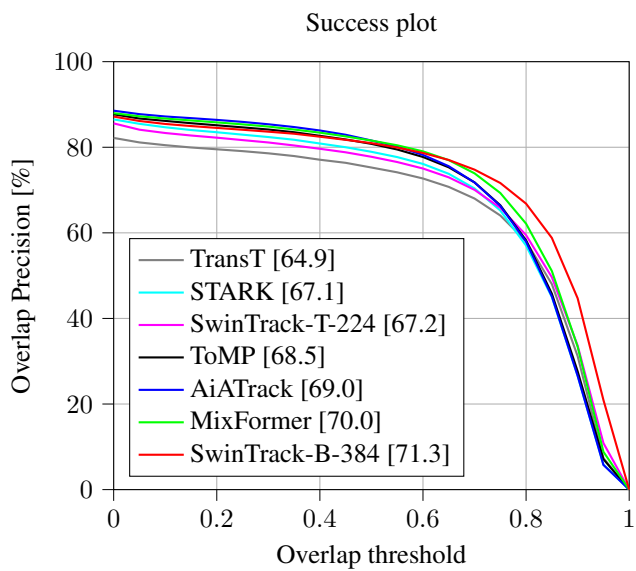
\begin{figure}
     \caption{Comparison to the state-of-the-art trackers on LaSOT~\cite{fan2019lasot} Test set using success (SUC) AUC score.}
    \label{success_fig}
     \centering
\begin{tikzpicture}

\definecolor{cyan}{RGB}{0,255,255}
\definecolor{darkgray176}{RGB}{176,176,176}
\definecolor{magenta}{RGB}{255,0,255}

\begin{axis}[
legend cell align={left},
legend style={fill opacity=1.0, draw opacity=1, text opacity=1, at={(0.03,0.03)}, anchor=south west},
tick align=outside,
tick pos=left,
title={Success plot},
x grid style={darkgray176},
xlabel={Overlap threshold},
xmajorgrids,
xmin=0, xmax=1,
xtick style={color=black},
y grid style={darkgray176},
ylabel={Overlap Precision [\%]},
ymajorgrids,
ymin=0, ymax=100,
ytick style={color=black}
]
\addplot [thick, gray]
table {%
0 82.1811447143555
0.0500000007450581 81.130126953125
0.100000001490116 80.4880905151367
0.150000005960464 79.9483261108398
0.200000002980232 79.524528503418
0.25 79.0952606201172
0.300000011920929 78.5835418701172
0.349999994039536 77.9458847045898
0.400000005960464 77.0856170654297
0.449999988079071 76.3620223999023
0.5 75.2638168334961
0.550000011920929 74.1415328979492
0.600000023841858 72.6995391845703
0.649999976158142 70.7366256713867
0.699999988079071 68.0166931152344
0.75 64.0507049560547
0.800000011920929 58.0414237976074
0.850000023841858 48.0121994018555
0.899999976158142 31.1894454956055
0.949999988079071 7.37785196304321
1 0
};
\addlegendentry{TransT [64.9]}
\addplot [thick, cyan]
table {%
0 86.5059585571289
0.0500000007450581 85.4760131835938
0.100000001490116 84.6729583740234
0.150000005960464 84.0289077758789
0.200000002980232 83.5075607299805
0.25 82.9430541992188
0.300000011920929 82.3839340209961
0.349999994039536 81.7521133422852
0.400000005960464 80.8750228881836
0.449999988079071 80.0048522949219
0.5 78.9352111816406
0.550000011920929 77.7029876708984
0.600000023841858 76.1019973754883
0.649999976158142 73.7927780151367
0.699999988079071 70.3733291625977
0.75 65.2167816162109
0.800000011920929 57.0196075439453
0.850000023841858 44.9836959838867
0.899999976158142 26.5942611694336
0.949999988079071 5.85685920715332
1 0
};
\addlegendentry{STARK [67.1]}
\addplot [thick, magenta]
table {%
0 85.5778732299805
0.0500000007450581 84.1035537719727
0.100000001490116 83.3299713134766
0.150000005960464 82.7461700439453
0.200000002980232 82.2377166748047
0.25 81.6727294921875
0.300000011920929 81.1188430786133
0.349999994039536 80.4158172607422
0.400000005960464 79.6402816772461
0.449999988079071 78.8200378417969
0.5 77.7748641967773
0.550000011920929 76.5463027954102
0.600000023841858 75.0328216552734
0.649999976158142 72.9672393798828
0.699999988079071 70.0464553833008
0.75 65.843864440918
0.800000011920929 59.4688186645508
0.850000023841858 49.6399803161621
0.899999976158142 33.6708564758301
0.949999988079071 10.8771266937256
1 0
};
\addlegendentry{SwinTrack-T-224 [67.2]}
\addplot [thick, black]
table {%
0 87.6947326660156
0.0500000007450581 86.7127380371094
0.100000001490116 86.1569366455078
0.150000005960464 85.6223678588867
0.200000002980232 85.134162902832
0.25 84.6478958129883
0.300000011920929 84.1479339599609
0.349999994039536 83.5117721557617
0.400000005960464 82.6738510131836
0.449999988079071 81.8229904174805
0.5 80.7913818359375
0.550000011920929 79.4765777587891
0.600000023841858 77.7239761352539
0.649999976158142 75.3065414428711
0.699999988079071 71.794792175293
0.75 66.445671081543
0.800000011920929 58.2587738037109
0.850000023841858 45.7767219543457
0.899999976158142 27.6395320892334
0.949999988079071 7.11588954925537
1 0
};
\addlegendentry{ToMP [68.5]}
\addplot [thick, blue]
table {%
0 88.5451202392578
0.0500000007450581 87.7382736206055
0.100000001490116 87.1921768188477
0.150000005960464 86.805793762207
0.200000002980232 86.3960647583008
0.25 85.9280548095703
0.300000011920929 85.3713989257812
0.349999994039536 84.7208786010742
0.400000005960464 83.9160537719727
0.449999988079071 82.9192733764648
0.5 81.5744247436523
0.550000011920929 80.1431274414062
0.600000023841858 78.2123794555664
0.649999976158142 75.5701751708984
0.699999988079071 71.7714157104492
0.75 66.3074264526367
0.800000011920929 57.933349609375
0.850000023841858 45.1459999084473
0.899999976158142 26.581958770752
0.949999988079071 5.77545738220215
1 0
};
\addlegendentry{AiATrack [69.0]}
\addplot [thick, green]
table {%
0 87.9642944335938
0.0500000007450581 87.2290115356445
0.100000001490116 86.7096862792969
0.150000005960464 86.2598266601562
0.200000002980232 85.8123626708984
0.25 85.3485641479492
0.300000011920929 84.8339157104492
0.349999994039536 84.1455917358398
0.400000005960464 83.4091491699219
0.449999988079071 82.5364990234375
0.5 81.5589218139648
0.550000011920929 80.4591064453125
0.600000023841858 79.0383453369141
0.649999976158142 76.987548828125
0.699999988079071 73.9069366455078
0.75 69.3097839355469
0.800000011920929 62.1433563232422
0.850000023841858 51.0754699707031
0.899999976158142 33.3884429931641
0.949999988079071 8.84209823608398
1 0
};
\addlegendentry{MixFormer [70.0]}
\addplot [thick, red]
table {%
0 87.1718978881836
0.0500000007450581 86.1381683349609
0.100000001490116 85.4462356567383
0.150000005960464 84.9568328857422
0.200000002980232 84.5217132568359
0.25 84.090202331543
0.300000011920929 83.642204284668
0.349999994039536 83.1625366210938
0.400000005960464 82.4673614501953
0.449999988079071 81.7340927124023
0.5 80.9639892578125
0.550000011920929 79.9329605102539
0.600000023841858 78.6637268066406
0.649999976158142 77.0235595703125
0.699999988079071 74.7673721313477
0.75 71.6431579589844
0.800000011920929 66.8184432983398
0.850000023841858 58.8123321533203
0.899999976158142 44.7114067077637
0.949999988079071 20.9500865936279
1 0
};
\addlegendentry{SwinTrack-B-384 [71.3]}
\end{axis}

\end{tikzpicture}
\end{figure}
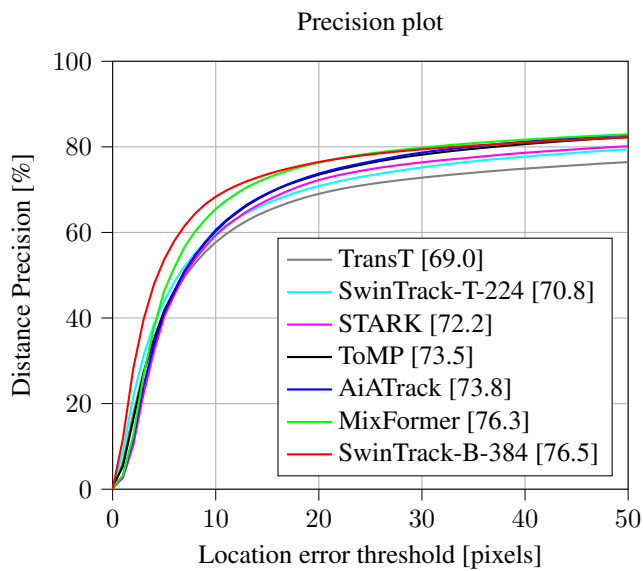
\begin{figure}
     \centering
     \caption{Comparison to the state-of-the-art trackers on LaSOT~\cite{fan2019lasot} Test set using precision (PRE) AUC score.}
    \label{precision_fig}
     \centering
\begin{tikzpicture}

\definecolor{cyan}{RGB}{0,255,255}
\definecolor{darkgray176}{RGB}{176,176,176}
\definecolor{magenta}{RGB}{255,0,255}

\begin{axis}[
legend cell align={left},
legend style={fill opacity=1.0, draw opacity=1, text opacity=1, at={(0.97,0.03)}, anchor=south east},
tick align=outside,
tick pos=left,
title={Precision plot},
x grid style={darkgray176},
xlabel={Location error threshold [pixels]},
xmajorgrids,
xmin=0, xmax=50,
xtick style={color=black},
y grid style={darkgray176},
ylabel={Distance Precision [\%]},
ymajorgrids,
ymin=0, ymax=100,
ytick style={color=black}
]
\addplot [thick, gray]
table {%
0 0.0504663623869419
1 6.35006046295166
2 17.6777515411377
3 27.51123046875
4 35.1705627441406
5 41.1083030700684
6 45.8147506713867
7 49.6967582702637
8 52.8312377929688
9 55.4575309753418
10 57.713550567627
11 59.6219940185547
12 61.252628326416
13 62.6995315551758
14 63.9634971618652
15 65.0405883789062
16 65.9823913574219
17 66.8850479125977
18 67.6832656860352
19 68.3992691040039
20 69.0166091918945
21 69.5805282592773
22 70.0909881591797
23 70.5376739501953
24 70.92724609375
25 71.3039703369141
26 71.6530914306641
27 71.9626617431641
28 72.2531585693359
29 72.5363845825195
30 72.7947082519531
31 73.0490875244141
32 73.2917861938477
33 73.5275650024414
34 73.7428741455078
35 73.9558868408203
36 74.1648101806641
37 74.3626861572266
38 74.5630416870117
39 74.7482299804688
40 74.9167938232422
41 75.0783462524414
42 75.2441482543945
43 75.4030456542969
44 75.5633392333984
45 75.7098388671875
46 75.8620147705078
47 76.0291213989258
48 76.1782684326172
49 76.3012466430664
50 76.4280395507812
};
\addlegendentry{TransT [69.0]}
\addplot [thick, cyan]
table {%
0 0.0504663623869419
1 8.82082366943359
2 21.6383800506592
3 31.2220611572266
4 38.507209777832
5 44.0193634033203
6 48.4878005981445
7 52.1530570983887
8 55.1679916381836
9 57.7008743286133
10 59.8693313598633
11 61.6847267150879
12 63.2172889709473
13 64.5631484985352
14 65.7556457519531
15 66.8354949951172
16 67.7942886352539
17 68.6865234375
18 69.4827346801758
19 70.1895065307617
20 70.8372116088867
21 71.4245758056641
22 71.9685516357422
23 72.4641418457031
24 72.9245529174805
25 73.3448638916016
26 73.756721496582
27 74.1450805664062
28 74.5152206420898
29 74.8672561645508
30 75.1938400268555
31 75.5105285644531
32 75.8148727416992
33 76.0864944458008
34 76.3699798583984
35 76.6304931640625
36 76.8664627075195
37 77.1056671142578
38 77.3082580566406
39 77.5058670043945
40 77.7110137939453
41 77.9050598144531
42 78.0927810668945
43 78.2720947265625
44 78.4523620605469
45 78.6244049072266
46 78.8016357421875
47 78.9714889526367
48 79.127571105957
49 79.2715682983398
50 79.4226837158203
};
\addlegendentry{SwinTrack-T-224 [70.8]}
\addplot [thick, magenta]
table {%
0 0.264687269926071
1 2.66286325454712
2 10.2780637741089
3 22.2662029266357
4 32.0875625610352
5 40.306453704834
6 45.5220031738281
7 50.2590065002441
8 53.7944679260254
9 56.7636795043945
10 59.3387718200684
11 61.4541282653809
12 63.2197494506836
13 64.878547668457
14 66.2517166137695
15 67.5147247314453
16 68.6433563232422
17 69.7103118896484
18 70.6519622802734
19 71.5002212524414
20 72.2395172119141
21 72.8751068115234
22 73.4193725585938
23 73.9048309326172
24 74.3532180786133
25 74.7707824707031
26 75.1520767211914
27 75.4855575561523
28 75.8070373535156
29 76.0983047485352
30 76.3693008422852
31 76.6468734741211
32 76.8952102661133
33 77.1156463623047
34 77.3304214477539
35 77.5635528564453
36 77.7762603759766
37 78.005615234375
38 78.2254409790039
39 78.4276809692383
40 78.6147689819336
41 78.7892303466797
42 78.9567642211914
43 79.121940612793
44 79.2837829589844
45 79.4345550537109
46 79.5879745483398
47 79.735107421875
48 79.8767013549805
49 80.0132751464844
50 80.1515197753906
};
\addlegendentry{STARK [72.2]}
\addplot [thick, black]
table {%
0 0.505607068538666
1 5.36885499954224
2 16.2277641296387
3 27.1090545654297
4 34.9986419677734
5 41.7345771789551
6 46.4616050720215
7 51.0711898803711
8 54.6231918334961
9 57.6373710632324
10 60.3525047302246
11 62.6116561889648
12 64.4953079223633
13 66.2497940063477
14 67.6984405517578
15 69.0177154541016
16 70.1415863037109
17 71.1519165039062
18 72.0068206787109
19 72.7753143310547
20 73.5076751708984
21 74.1445083618164
22 74.7381820678711
23 75.2673492431641
24 75.7610168457031
25 76.259391784668
26 76.704475402832
27 77.1144180297852
28 77.5067825317383
29 77.8565902709961
30 78.1669769287109
31 78.4766845703125
32 78.7707901000977
33 79.0440521240234
34 79.3026885986328
35 79.5458145141602
36 79.7714309692383
37 80.0000686645508
38 80.2222671508789
39 80.4377670288086
40 80.6393356323242
41 80.8413619995117
42 81.0181045532227
43 81.2052612304688
44 81.3865509033203
45 81.546989440918
46 81.7103118896484
47 81.8591690063477
48 82.0022811889648
49 82.1379089355469
50 82.2716445922852
};
\addlegendentry{ToMP [73.5]}
\addplot [thick, blue]
table {%
0 0.264570116996765
1 3.14332580566406
2 11.717978477478
3 24.13014793396
4 33.6136779785156
5 41.3100738525391
6 46.273509979248
7 51.0369529724121
8 54.6928596496582
9 57.8181037902832
10 60.5801086425781
11 62.8392524719238
12 64.7144622802734
13 66.474494934082
14 67.8798522949219
15 69.1258316040039
16 70.2530364990234
17 71.2731475830078
18 72.1784057617188
19 73.0053482055664
20 73.7553100585938
21 74.4172134399414
22 75.0283813476562
23 75.5589065551758
24 76.066291809082
25 76.5566329956055
26 77.0350875854492
27 77.4806365966797
28 77.9204177856445
29 78.3084869384766
30 78.6523590087891
31 78.9783325195312
32 79.2682647705078
33 79.5588989257812
34 79.8181076049805
35 80.0702133178711
36 80.3110885620117
37 80.5658874511719
38 80.7897186279297
39 81.0054397583008
40 81.2042617797852
41 81.4077529907227
42 81.5974884033203
43 81.7708206176758
44 81.9353561401367
45 82.102912902832
46 82.2528228759766
47 82.3956909179688
48 82.5403442382812
49 82.6825790405273
50 82.8237380981445
};
\addlegendentry{AiATrack [73.8]}
\addplot [thick, green]
table {%
0 0.304750382900238
1 3.23846530914307
2 12.5862970352173
3 26.6148262023926
4 37.6737747192383
5 46.376880645752
6 51.8484420776367
7 56.7283515930176
8 60.223503112793
9 63.0621147155762
10 65.4960861206055
11 67.4364929199219
12 69.0585098266602
13 70.5485000610352
14 71.7113876342773
15 72.7361907958984
16 73.6279449462891
17 74.4316482543945
18 75.1371002197266
19 75.7486801147461
20 76.3344039916992
21 76.8381195068359
22 77.33544921875
23 77.7434387207031
24 78.1199722290039
25 78.4653930664062
26 78.7747116088867
27 79.0489730834961
28 79.3204879760742
29 79.5672836303711
30 79.8155517578125
31 80.0505752563477
32 80.2650985717773
33 80.4735488891602
34 80.6727676391602
35 80.8591613769531
36 81.0416107177734
37 81.211784362793
38 81.3712310791016
39 81.5326538085938
40 81.6760635375977
41 81.8192443847656
42 81.9595108032227
43 82.1002044677734
44 82.2351379394531
45 82.3682479858398
46 82.4833831787109
47 82.5873336791992
48 82.6947784423828
49 82.8041152954102
50 82.9195098876953
};
\addlegendentry{MixFormer [76.3]}
\addplot [thick, red]
table {%
0 0.0504663623869419
1 11.8184518814087
2 28.1475982666016
3 39.6667289733887
4 47.7942314147949
5 53.6860885620117
6 58.1003837585449
7 61.6025276184082
8 64.4151992797852
9 66.6277313232422
10 68.3537902832031
11 69.7637023925781
12 70.930549621582
13 71.9451446533203
14 72.8109664916992
15 73.5948333740234
16 74.2950057983398
17 74.9093780517578
18 75.4595718383789
19 75.9770965576172
20 76.4579238891602
21 76.8763275146484
22 77.2583541870117
23 77.6085052490234
24 77.935188293457
25 78.2289581298828
26 78.5010604858398
27 78.7525863647461
28 78.9847564697266
29 79.2096710205078
30 79.4118957519531
31 79.6002960205078
32 79.7815246582031
33 79.9609222412109
34 80.1327056884766
35 80.3031997680664
36 80.45068359375
37 80.5997009277344
38 80.7397308349609
39 80.8952789306641
40 81.0318756103516
41 81.1686553955078
42 81.2992172241211
43 81.4214248657227
44 81.5451126098633
45 81.6689224243164
46 81.7865219116211
47 81.9021072387695
48 82.0111389160156
49 82.1211853027344
50 82.2361602783203
};
\addlegendentry{SwinTrack-B-384 [76.5]}
\end{axis}

\end{tikzpicture}
\end{figure}

\afterpage{
\begin{figure}
\caption{Heatmap visualization of the failure cases. The organizational form is the same as Fig.~ \ref{heatmap_viz}. The sequences and challenges involved: badminton-3 in LaSOT$_{\text{ext}}$ (MB, SV, FOC, FM, OV, LR, ARC), skatingshoe-2 in LaSOT$_{\text{ext}}$ (POC, MB, ROT, BC, SV, FOC, FM, LR, ARC) and conduction1 (non-semantic target) in VOT-STB2022.\protect\footnotemark}
\label{heatmap_fail_viz}
\centering
\bgroup
\setlength{\tabcolsep}{1pt}
\begin{tabular}{cccc}
\subfloat{\includegraphics[width = 0.15\textwidth]{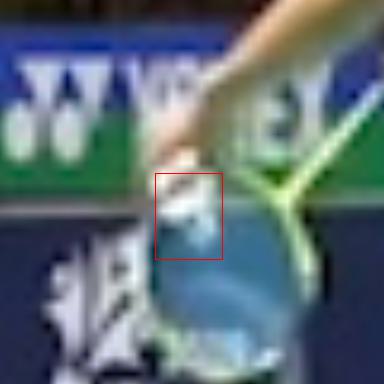}} &
\subfloat{\includegraphics[width = 0.15\textwidth]{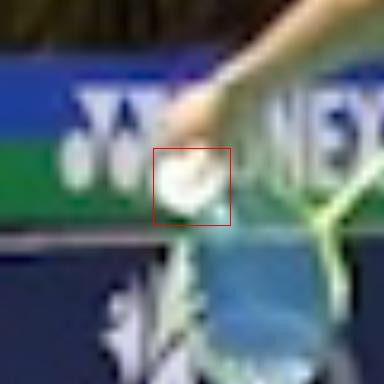}} &
\subfloat{\includegraphics[width = 0.15\textwidth]{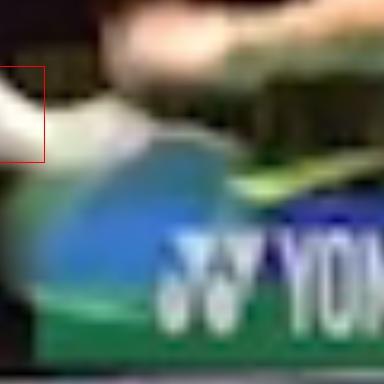}} &
\subfloat{\includegraphics[width = 0.15\textwidth]{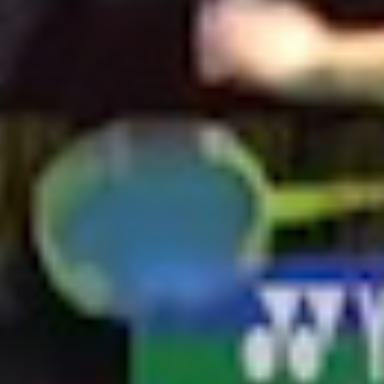}}\\
[-16pt]
\subfloat{\includegraphics[width = 0.15\textwidth]{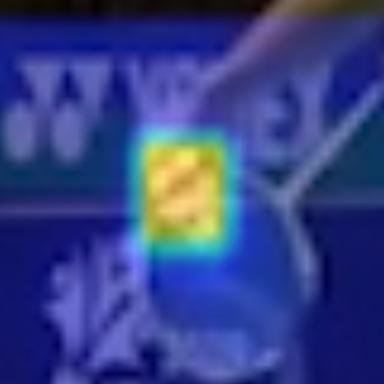}} &
\subfloat{\includegraphics[width = 0.15\textwidth]{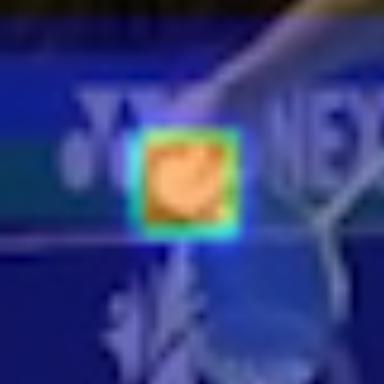}} &
\subfloat{\includegraphics[width = 0.15\textwidth]{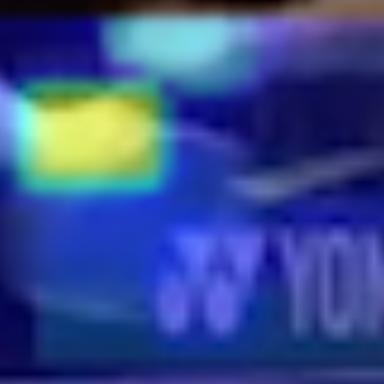}} &
\subfloat{\includegraphics[width = 0.15\textwidth]{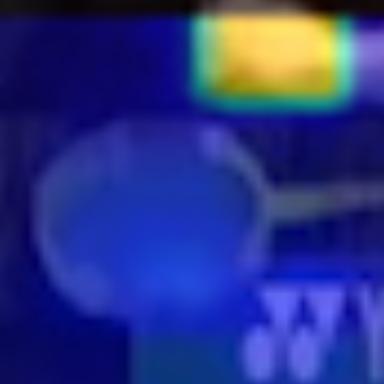}}\\
[-14pt]
\subfloat{\includegraphics[width = 0.15\textwidth]{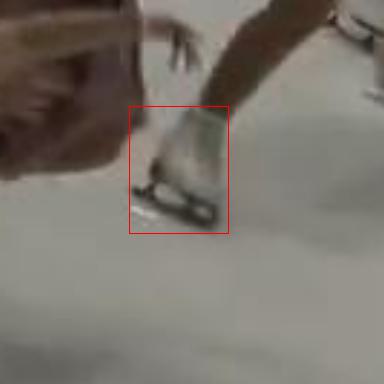}} &
\subfloat{\includegraphics[width = 0.15\textwidth]{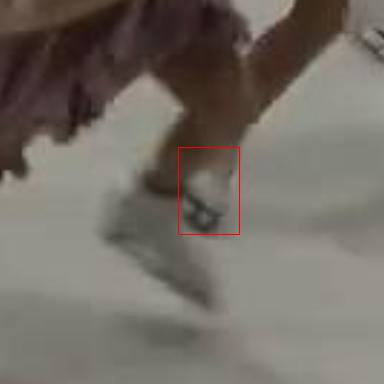}} &
\subfloat{\includegraphics[width = 0.15\textwidth]{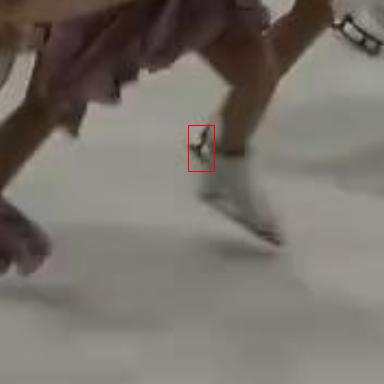}} &
\subfloat{\includegraphics[width = 0.15\textwidth]{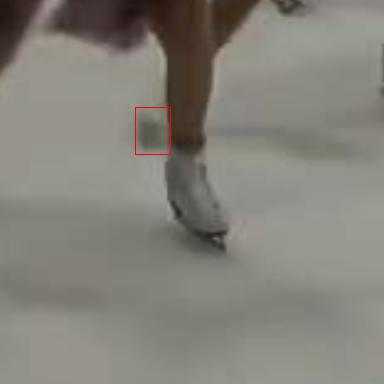}}\\
[-16pt]
\subfloat{\includegraphics[width = 0.15\textwidth]{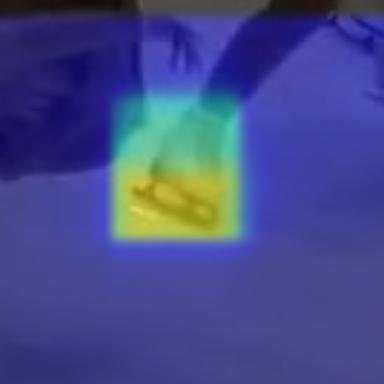}} &
\subfloat{\includegraphics[width = 0.15\textwidth]{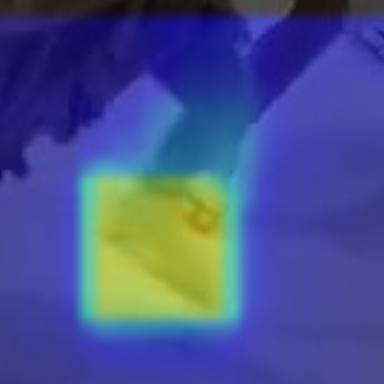}} &
\subfloat{\includegraphics[width = 0.15\textwidth]{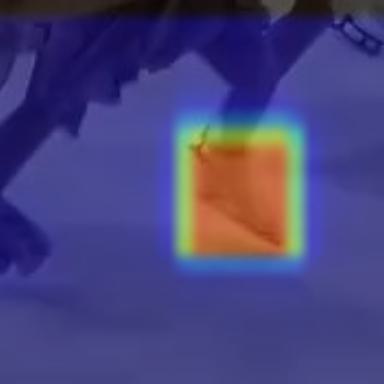}} &
\subfloat{\includegraphics[width = 0.15\textwidth]{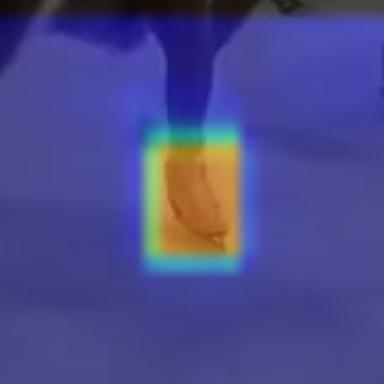}}\\
[-14pt]
\subfloat{\includegraphics[width = 0.15\textwidth]{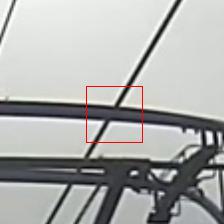}} &
\subfloat{\includegraphics[width = 0.15\textwidth]{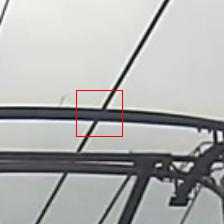}} &
\subfloat{\includegraphics[width = 0.15\textwidth]{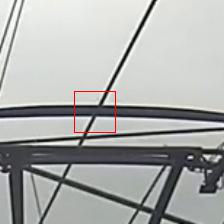}} &
\subfloat{\includegraphics[width = 0.15\textwidth]{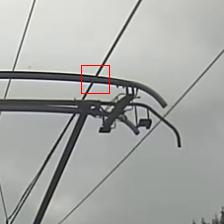}}\\
[-16pt]
\subfloat{\includegraphics[width = 0.15\textwidth]{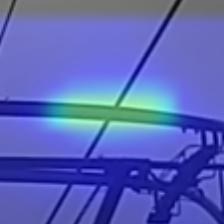}} &
\subfloat{\includegraphics[width = 0.15\textwidth]{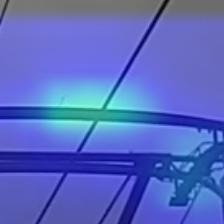}} &
\subfloat{\includegraphics[width = 0.15\textwidth]{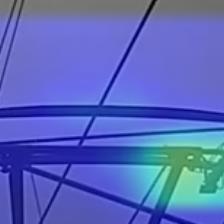}} &
\subfloat{\includegraphics[width = 0.15\textwidth]{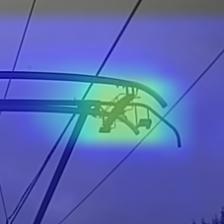}}
\end{tabular}
\egroup
\end{figure}
\footnotetext{IV: Illumination Variation, POC: Partial Occlusion, DEF: Deformation, MB: Motion Blur, CM: Camera Motion, ROT: Rotation, BC: Background Clutter, VC: Viewpoint Change, SV: Scale Variation, FOC: Full Occlusion, FM: Fast Motion, OV: Out-of-View, LR: Low Resolution, ARC: Aspect Ration Change}
}

\end{document}